\documentclass[preprint,12pt]{elsarticle}

\usepackage{amssymb}
\usepackage{amsmath}
\usepackage{wrapfig}
\usepackage{caption}

\usepackage{algorithm} 
\usepackage{algpseudocode} 

\makeatletter
\newenvironment{breakablealgorithm}
  {
     \refstepcounter{algorithm}
     \hrule height.8pt depth0pt \kern2pt
     \renewcommand{\caption}[2][\relax]{
       {\raggedright\textbf{\fname@algorithm~\thealgorithm} ##2\par}%
       \ifx\relax##1\relax 
         \addcontentsline{loa}{algorithm}{\protect\numberline{\thealgorithm}##2}%
       \else 
         \addcontentsline{loa}{algorithm}{\protect\numberline{\thealgorithm}##1}%
       \fi
       \kern2pt\hrule\kern2pt
     }
  }{
     \kern2pt\hrule\relax
  }
\makeatother

\usepackage{pdflscape}

\usepackage[hidelinks]{hyperref}

\journal{}

\begin{document}

\begin{frontmatter}



\title{From Data to Action: Accelerating Refinery Optimization with AI}


\author[BME]{Dániel Pfeifer} 
\author[BME]{Ábrahám Papp} 
\author[MOL]{Tibor Bernáth} 
\author[MOL]{Tamás Zoltán Varga} 
\author[MOL]{Márk Czifra} 
\author[BME]{Botond Szilágyi} 
\author[BME]{Edith Alice Kovács} 

\affiliation[BME]{organization={Budapest University of Technology and Economics},
            addressline={Műegyetem rkp. 3.}, 
            city={Budapest},
            postcode={1111}, 
            country={Hungary}}

\affiliation[MOL]{organization={MOL Group},
            addressline={Dombóvári út 28.}, 
            city={Budapest},
            postcode={1117}, 
            country={Hungary}}

\begin{abstract}
Nowadays refinery optimization utilizes sheer amounts of data, which can be handled with modern Linear Programming (LP) software, but the interpreting and applying the results remains challenging. Large petrochemical companies use massive models, with hundreds of thousands of input matrix elements. The LP solution is mathematically correct, but simplifications are made in the model, and data supply errors may occur. Therefore, further insight is needed to trust the results.

The novelty of the paper consists of enhancing the LP solver with a memory, which makes possible to find potential errors in the input data, finding unusual relations in the data on different decision levels, by introducing an anomaly detection tool. Analyzing the potential "anomalies" in the input and output of LP, the subsequent LP solutions can be improved.

The Anomaly Detection tool proposed comprises three methodologies. A transformed version of the popular ECOD methodology is applied. New methods are proposed to handle high-dimensional data: choosing the most informative pairs. Then, this is used alongside two 2D Anomaly Detection algorithms, revealing several business opportunities and data supply errors in the MOL refinery scheduling and planning architecture.

The proposed tool learns connections in past data and runs very efficiently on present data revealing unusual connections sorted by their importance. Up to this step, optimization and Anomaly Detection happened automatically, and thereafter, decision makers may take a look at the most relevant anomalies found, and decide how to further improve the subsequent LPs. The tool also has periodic retraining built in. 
\end{abstract}


\begin{highlights}
\item Enhancing LP-based decision making with historical data.
\item Detecting anomalies in LP input and output data improves the quality of LP solutions.
\item Unusual data relations across decision levels reveal business opportunities.
\item Fast and efficient methods for multivariate anomaly detection in big data, based on finding informative margins.
\item Revealing business opportunities and understanding key bottlenecks by analyzing types of anomalies.
\end{highlights}

\begin{keyword}
Anomaly Detection \sep
Big Data \sep
Linear Programming \sep
Marginal Values \sep
Business Opportunities
\footnote{e-mail addresses: Dániel Pfeifer: pfeiferd@math.bme.hu, Ábrahám Papp: abrapapp@math.bme.hu, Tibor Bernáth: tbernath@mol.hu, Tamás Zoltán Varga: tvarga@mol.hu, Márk Czifra: maczifra@mol.hu, Botond Szilágyi: szilagyi.botond@vbk.bme.hu, Edith Alice Kovács: kovacsea@math.bme.hu}


\end{keyword}

\end{frontmatter}



\section{Introduction}\label{intro}

Modern technology applications generate vast amounts of data. Discovering hidden patterns within existing data and leveraging them for business insights is becoming increasingly important.

In general, Big Data refers to large and complex datasets that are often used for predictive analytics. It is a term that relates to massive, heterogeneous, and often unstructured digital data that are difficult to handle with traditional data management tools and techniques \cite{rodriguez2016general}.

The data analysis-based technologies that various organizations and companies have recently implemented are called Business intelligence. In M. Hamzehi et al. \cite{hamzehi2022business}, the data set used is a data set related to the sales system of a pharmaceutical company, which enters the data into a data warehouse online. The paper aims to provide an efficient model for optimizing the products sales system in a pharmaceutical company using clustering methods based on machine learning indicators and algorithms.
Rath et al. \cite{rath2021realization} introduced a structure to build a Business Intelligence framework. They show that machine learning approaches play an important role in the tasks of Business Intelligence in large commercial organizations.

Another type of investigation is presented in a recent paper by F. Ridzuan et al. \cite{ridzuan2022diagnostic} related to the problem of outliers. It includes recommendations to improve the quality of data and data collection systems are provided by finding several factors that greatly influence the presence of outliers.

Machine learning approaches are intensively applied to address real-world problems in the fourth industrial revolution. These kinds of approaches are now an integral part of the operations of most oil and gas companies, which they use to translate datasets, including large volumes of real-time information into actionable insights. A good overview of these methods is given in \cite{larranaga2018industrial} and \cite{carou2022machine}. Aspects related to big data in this field are discussed in \cite{klipa2022big}.

A clear literature overview of the methodologies for addressing scheduling, planning, and supply chain management of oil refinery operations can be found in \cite{shah2011petroleum}. A paper by Grossmann \cite{grossmann2005enterprise} establishes enterprise-wide optimization as the framework for integrating supply, manufacturing, and distribution decisions across multiple levels and sites. Another perspective on surveying the history and current state of AI and machine learning applications in chemical engineering can be found in \cite{venkatasubramanian2019promise}.

The research presented in this paper is motivated by the problems which occur during the application of an efficient optimization solver in a Big Data environment.

Linear Programming (LP) models have been a staple in Refinery Planning and Optimization since the late 20th century. These large mathematical models take hundreds of thousands of inputs and produce hundreds of thousands of outputs, to find the most economical way to operate one or more refineries, petrochemical plants, their feedstock supply, market demands all while considering every relevant logistic or operational constraint.

In the past decades, these models have become necessary, as finding the optimum and maximizing profits have become a must among shrinking margins, rising energy prices, volatile external environment and ever stricter regulatory pressure on the industry. An even more developed idea in a recent study to apply an AI safety system that operates in parallel with operational AI, continuously supervising system safety behavior can be found in \cite{thakur2026artificial}.

MOL Group is a Central European integrated oil company, that has also been using an LP optimizer, AspenTech PIMS (Process Industry Modelling System) \cite{aspenunified} for more than 30 years to optimize their monthly plans and evaluate the refinery economics. During this time a wide range of supporting tools were developed internally with the sole purpose of helping to understand the result of the LP optimization. Due to the sheer amount of data and the complex technological and optimization knowledge needed to interpret the results it remains a daunting task.

While LP gives a mathematically optimal solution it is the optimization experts' responsibility together with refinery scheduling, logistics and other disciplines to make sure that these plans make sense not just mathematically but economically and operationally as well. Analyzing the entirety of the input \& output data is not possible as of now, since the size and complexity of the problem is in the world of big data. So far key process and economic indicators were defined and monitored, however everchanging external factors such as Covid, NatGas price volatility and other side effects of the RUS-UA war can fundamentally change main drivers of optimization. To tackle these obstacles, AI solutions may be used to help our understanding, find business opportunities and maximize margins. 

Anomaly Detection tools are being implemented that compare the inputs \& outputs of new plans to old ones which have already been checked by all stakeholders and accepted by the management. This study will focus on three new Anomaly Detection methodologies that have been fine-tuned by applying them to real data from monthly plans and illustrate how LP and Anomaly Detection can go hand in hand to enhance each other's, and the company’s performance. 

The remainder of the paper is the following. In Section \ref{section:refplan}, we introduce the Linear Programming methodology used to solve Refinery Optimization problems. We detail how Marginal and Incremental Values help experts find business opportunities, and introduce the LP software used. In Section \ref{section:goal_anomdet}, we introduce Anomaly Detection as a tool for finding such business opportunities. Through several real life examples, we show kinds of anomalies that may occur. Finally, we introduce the input data that the Anomaly Detection will use. In Section \ref{section:anomdet}, we detail the exact mathematical methodology of Anomaly Detection, and introduce two 2D algorithms and one 1D modified algorithm for detecting values or value pairs that do not conform the historical distribution. Additionally, we introduce a pair selection methodology that can vastly reduce the number of possible pairs to look through. In Section \ref{section:results}, we illustrate some real life 1D and 2D anomalies we have found with this methodology. Finally, Section \ref{section:summary} summarizes the paper.

\section{Refinery Planning \& Optimization}\label{section:refplan}

\subsection{Linear Programming methodology}

A central methodology of oil refinery planning is Linear Programming. An early formulation of an LP/NLP refinery planning model, covering processing units, intermediate streams, and blending relations with real industrial case studies can be found in \cite{pinto2000planning}. Here, we will introduce the primal and dual Linear Programming problems and show how they are related to each other.

Let $\mathbf{A} \in \mathbb{R}^{m \times n}$ be the technological matrix obtained from data supply containing known parameters (e.g. prices and costs, capacities, blending proportions etc.),
$\mathbf{x} \in \mathbb{R}^n$ the decision variables of the primal problem (e.g. amounts bought and sold, amounts produced, amounts transferred, etc.), $\mathbf{b} \in\mathbb{R}^m$ the constraint limits of the primal problem, (e.g. upper bounds of the available resources), and $\mathbf{c} \in\mathbb{R}^n$ the weights of the objective function.

The primal linear programming problem has the following form:
\begin{equation}
\begin{split}\label{primal}
\text{(P)} \quad 
\max_{\mathbf{x} \in \mathbb{R}^n} \quad & \mathbf{c}^{T} \mathbf{x}\\
\text{subject to} \quad 
& \mathbf{A x }\le \mathbf{b}, \\
& \mathbf{x} \ge \mathbf{0}
\end{split}
\end{equation}
In the primal problem we maximize a linear objective function, the solutions (decision variables) are the elements of $\mathbf{x}$ vector. Based on the primal problem (Formula \ref{primal}), we can formulate the dual problem in the following way:
\begin{equation}
\begin{split}
\text{(D)} \quad 
\min_{\mathbf{y} \in \mathbb{R}^m} \quad & \mathbf{b}^{T} \mathbf{y} \\
\text{subject to} \quad 
& \mathbf{A}^{T} \mathbf{y} \ge\mathbf{c}, \\
&\mathbf{y} \ge \mathbf{0}
\end{split}
\end{equation}
The decision variables of the dual problem are contained in $\mathbf{y}\in \mathbb{R}^m$.
Its components $y_i$ are the dual decision variables, one for each primal constraint.
By the strong duality theorem \cite[p.~146-155]{bertsimas1997introduction} if both problems have solutions, then the maximum of the primal objective and the minimum of dual objective functions are equal.

In the refinery case, there are more unknowns than equations, therefore $n > m$.

Marginal Values are an important concept in LP. Their values are generated in each LP run. They can be extracted for all constraints. Non-zero Marginal values appear in the cases where variables are non basis.

Marginal Values indicate what would be the increment of the objective (gross profit) function in case of an infinitesimal change on the given bound. The more the given bound constrains the profit generation the higher the Marginal Value will be. If the given bound is not constraining, the profit generation there is no (or $0$) MV on that constraint.

The Marginal Value of a row of constraints $\mathbf{a}_i^T\mathbf{x} \le b_i$ is

$$P_i = \frac{\partial}{\partial b_i} \max \{ \mathbf{c}^T \mathbf{x} | \mathbf{A}\mathbf{x}\le \mathbf{b}\}$$

From the dual problem, the Marginal Value of a column can also be defined. Since the dual problem has a transposed $\mathbf{A}$ matrix, each constraint in the dual problem can be interpreted as a column of the original problem. So the Marginal Value of a column $j$ is

$$D_j = \frac{\partial}{\partial c_j} \min \{ \mathbf{b}^T \mathbf{y} | \mathbf{A}^T\mathbf{y}\ge \mathbf{c}\}$$

A constraint $\mathbf{a}_i^T \mathbf{x} \le b_i$ is called basis, if $\mathbf{a}_i \hat{\mathbf{x}} = b_i$ holds where $\hat{\mathbf{x}}$ is the solution of the LP.

\subsection{Refinery Economics}

LP optimization is particularly valuable in the complex environment of refineries and petrochemical plants, where numerous variables and constraints must be managed simultaneously.

\vspace{3mm}
\textbf{Marginal Values (MV)}

Marginal Values are rates of change of the Objective Function with a change in the activity of a variable. As partial derivatives of the Objective Function – they are only valid at the increments around the optimal activity.

\begin{wrapfigure}{r}{0.4\textwidth}
  \vspace{-10mm}
  \begin{center}
    \includegraphics[width=0.38\textwidth]{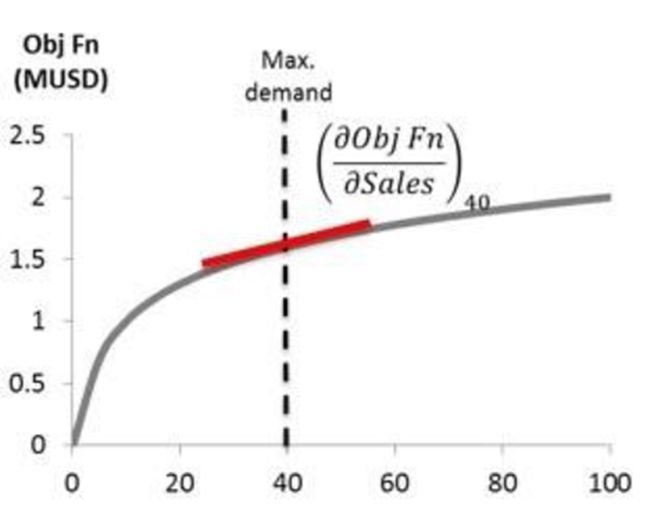}
  \caption{Example: How much we would earn (USD/t) on selling the 40001st ton? Marginal Values are only valid for the increments around the original $40\ 000$ tons, while BEV considers average for the next $40$ kt.}
  \end{center}
  \label{fig:MV}
  \vspace{-10mm}
\end{wrapfigure}

$D_j:$ Applies to column activities (variables), and represents rate of change in objective function as the bound (min or max) of a variable is increased.

$P_i:$ Applies to rows (equations), and represents the rate of change in the objective function as the constraint (min or max) of a row is decreased.

\vspace{3mm}
\textbf{Incremental Values (IV)}

Incremental Values are internal values of a material within the planning premises. For any active material and utility in the LP an incremental value is defined as well. These refer to the substitution value of the infinitesimal amount of the given product.

$D_j$ of a Basis Column: All basis columns will have a $D_j$ equal to zero because the cost of producing the next increment and its incremental product value balance each other and the slope is zero. If a column has an entry (price or cost) in the objective function row, the $D_j$ will be the difference between this entry and the incremental value of the column in a solution.

Break-even Value: A break-even activity is a value which does not alter a refiner’s economics. In other words, they do not lose (nor make) money in the process. The break-even value of an activity is the unit price (or cost) of the activity that preserves the original objective function of the refinery LP model before the specified changes have taken place. The break-even position provides a benchmark for negotiating a discount or premium of a transaction.

\subsection{Detecting Business Opportunities with Incremental and Marginal Values}

Marginal values (MV) are a key concept in LP, and support understanding the optimization drivers and result in a complex system. These values are generated in each LP run.

Incremental Values are obtained from Marginal Values. They can only be calculated for variables $x_i$ for which $c_i \neq 0$ (or in other words, variables that appear in the objective function). In this case, these variables represent some physically ready material that can be sold, and, as such, we are interested in how its sell price is assembled. Incremental Value is then equal to the buy costs of this product, plus its production costs, transfer costs, storage costs, etc., minus its Marginal Value. At every stage of production, this represents how valuable the given product is inside the refinery.

\begin{figure}[h]
    \centering
    \includegraphics[width=0.99\linewidth]{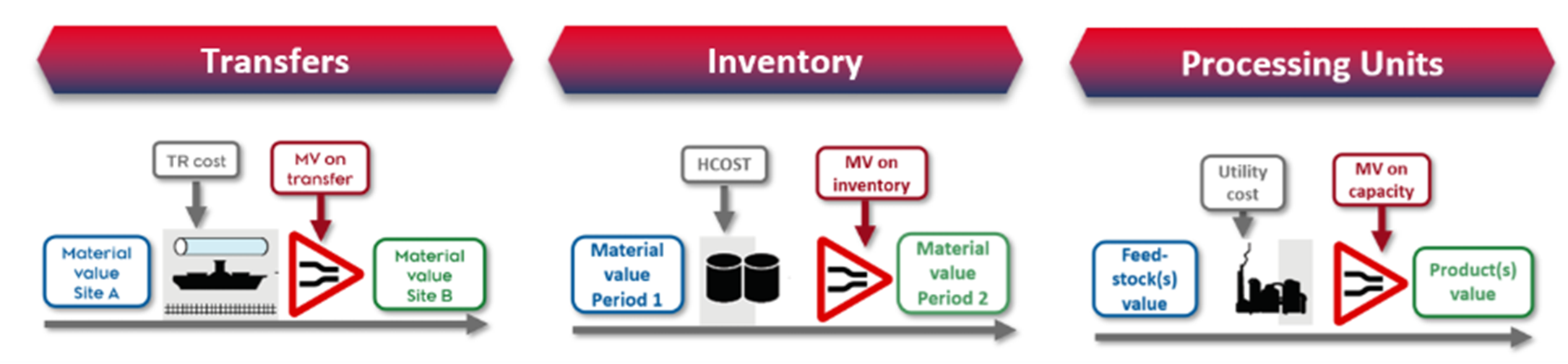}
    \caption{Simple representation of Marginal and Incremental Value dynamics. Any transformation in the LP (transfer between sites, inventorying between periods or a processing unit) can be described with the same methodology. Incremental and Marginal values are connected and can be calculated from each other.}
    \label{fig:MV_IV_dynamics}
\end{figure}

Marginal and Incremental Values are real treasures for an optimizer, and most of the time they are the key for understanding complex optimization dynamics. Therefore the MOL Group Optimization team developed multiple reports and concepts to understand the LP results through MVs:

\begin{itemize}
\item Reports for ranked MVs to capacities, sales/purchases, other constraining variables are especially useful to indicate the most important bottlenecks and optimization drivers, often revealing model stability and improper settings.
\item Graphs showing values of same materials across multiple periods and sites – explaining market sharing, transfer and inventory dynamics.
\item A simple diagrams explaining MV dynamics for transfers, inventory and unit utilization.
\item Graphical reports of unit feed and product MVs revealing bottleneck products and asymmetries in operation. 
\item Multi-case (direction) runs specifically focusing on MV dynamics of a given material and downstream inflection points.
\end{itemize}

\begin{figure}[h]
    \centering
    \includegraphics[width=0.99\linewidth]{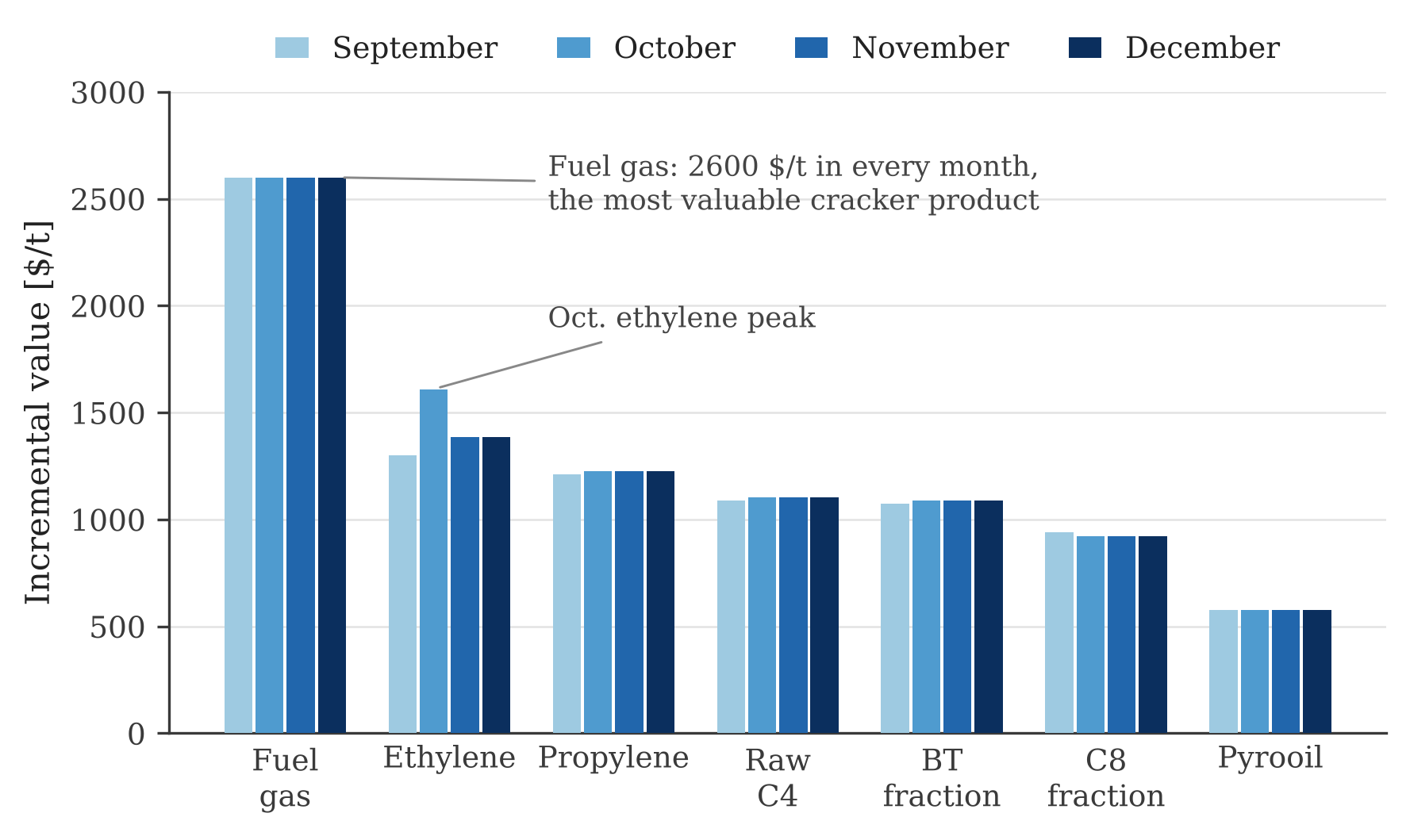}
    \caption{A product value diagram for one of MOL Group’s steam crackers, during the times of the historically high natural gas prices. In these months the historically lowest value Fuel gas had been the most valuable product of the Steam crackers. Cracker operation differences among the periods can be understood easier by observing Ethylene value differences among the periods.}
    \label{fig:steam_cracker_product_IVs}
\end{figure}

\begin{figure}[h]
    \centering
    \includegraphics[width=0.99\linewidth]{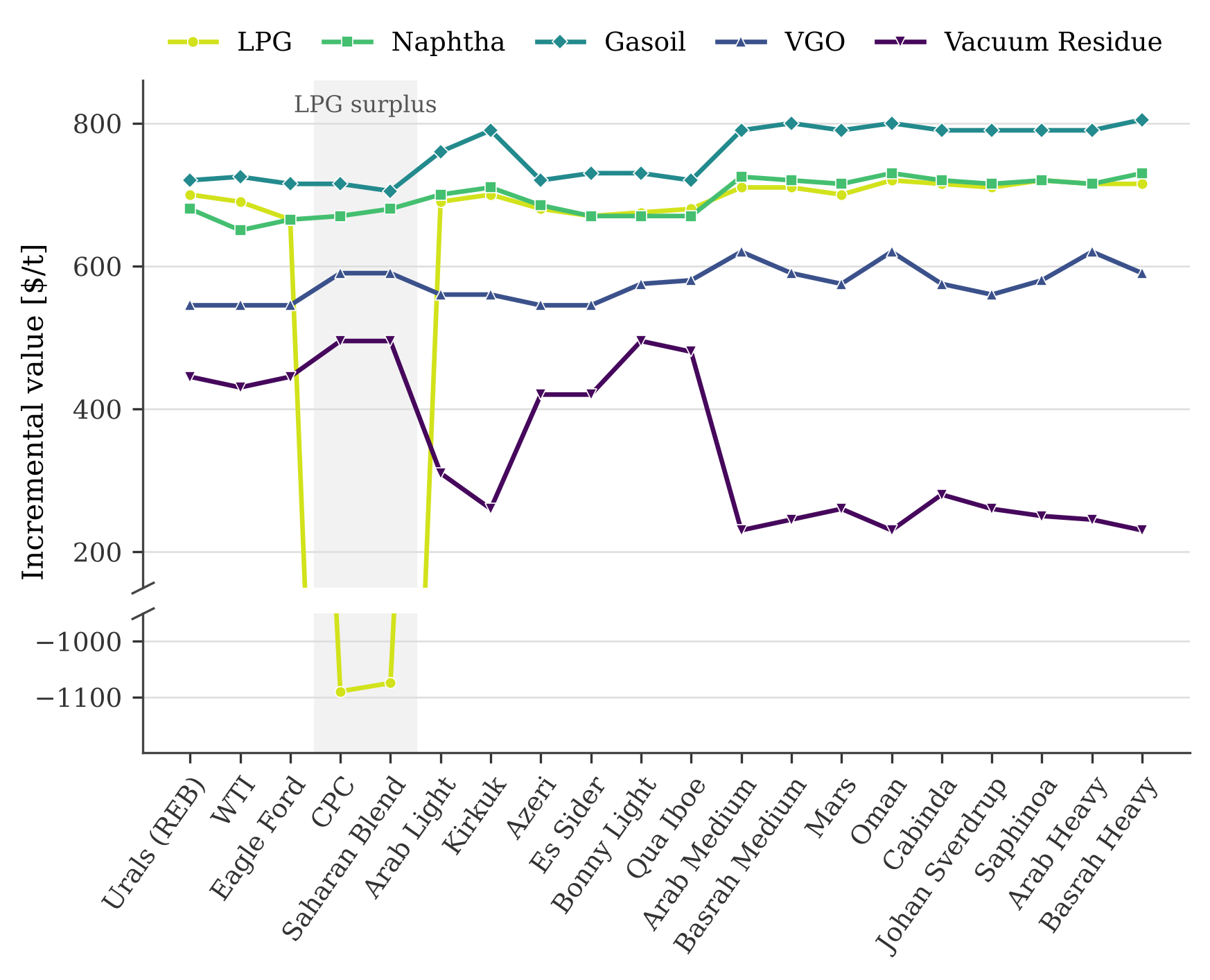}
    \caption{A product value diagram from multiple LP runs – in this case a crude evaluation. These graphs can reveal how different straight run products’ (direct products of crude distillation units) values change in case of different crudes processed, showing the product shortage (when high value is observed) or product surplus (low value observed – in extreme cases even negative ones)}
    \label{fig:product_IV}
\end{figure}

Although these reports and developments already give strong support for result analysis (see Figures \ref{fig:steam_cracker_product_IVs}, \ref{fig:product_IV}), the dashboards are only created from a fraction of the LP results; and all of them focus on a single LP run, with no memory and not using relationships of variables from historical optimization results.

This necessitates the introduction of a more sophisticated method that does have a memory and can use and connect historical information.

\subsection{Linear Programming software - AspenTech PIMS}

It is a flexible and easy to use productivity tool for economic planning in the process industries. PIMS (Process Industry Modelling System) \cite{aspenunified} employs Linear Programming technique to optimize the operation and economics of refineries, petrochemical plants, or other industry facilities.

The $\mathbf{A}$ matrix representing a PIMS model is a collection of linear equations. Each of these equations represents one aspect of the refinery or petrochemical plant operation, logistics or commercial constraint. It may be used for a wide variety of short term and strategic planning purposes, such as the evaluation of alternative feedstocks, optimization of product slates, evaluation of grass root opportunities or expansions, and many others. Although PIMS is called a linear programming system, it has many nonlinear elements.

Nowadays, the use of PIMS is common in the oil industry, as about $75\%$ of the world's refineries and about $60\%$ of petrochemical companies use it for production planning and scheduling \cite{aspenunified}. MOL Group has been using the PIMS software, by AspenTech since 1993.

\section{Goal of Anomaly Detection}\label{section:goal_anomdet}

\begin{figure}[h!]
    \centering
    \includegraphics[width=0.99\linewidth]{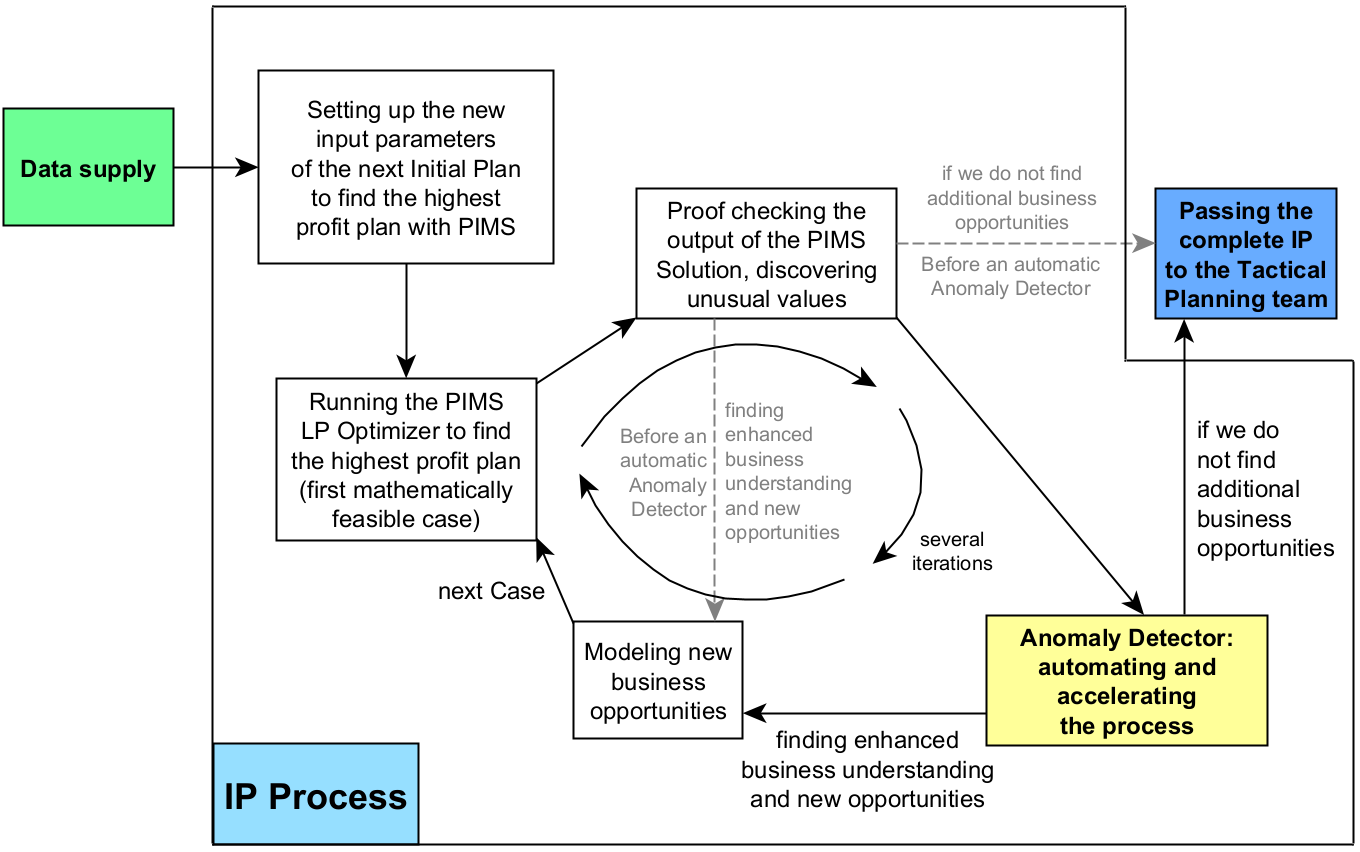}
    \caption{The Initial Plan (IP) process. It is the planning and optimization process in MOL Group, done on a monthly basis, to provide optimized plans for the whole Downstream business for the upcoming 3-4 months. The first step in the planning is the data supply, based on which operational and external (economic) data is updated in the model. Based on this “raw” data the first mathematically feasible case (C00) is created. This is followed by an iterative process, together with scheduling, logistic and other stakeholders the plan is finetuned to reflect the operational reality. The Anomaly Detector aims to support this iterative process.}
    \label{fig:IP_process}
\vspace{-5mm}
\end{figure}

\subsection{Real life challenges of refinery optimization}

MOL Group is an integrated oil and gas company, operating three refineries and two petrochemical sites in Central Europe in relative proximity ($\sim500$km), with the majority of the sites being landlocked. This makes logistics and inventory management key optimization drivers, which require multi-site, multi-period (3-4 month) models. For this, MOL Group is using one of the largest and most complex XPIMS models in the world. The current model matrix consists of $\sim 45\ 000$ rows, $\sim 60\ 000$ columns and over $600\ 000$ non-zero elements. The size of the mathematical model in itself represents a significant challenge regarding feasibility, stability and solving time.

On top of this the data management and business understanding needed for the optimization and economic calculations require years of experience in the industry and a deep understanding in the field of optimization with linear programming. 

The key challenges of data management is the data supply process, which is the first step of every planning cycle and includes dozens of stakeholders providing their corresponding data from external environment; as well as operational parameters and planned inventory levels to be set as drivers or constraints in the model. Due to the sheer amount of data in this process, mistakes (like missing or wrong values) and inconsistencies (illogical values) or discrepancies (e.g. MIN>MAX)) are unavoidable, and we need to handle them. This is a classic case of “garbage in garbage out”.

Another key challenge of optimization stems from the complexity of interpreting the outputs (results) of the model. We need to make sure the result is not just mathematically optimal, but it is operationally viable as well, and makes sense economically. Interpreting results, understanding relations and the economic drivers require deep domain expertise, analytical thinking, and years of experience.

In refinery optimization, monthly plans are being created, which means one month is one period. What this means is when the monthly balance is being calculated PIMS only sees the beginning and end point of the month, which in this case is the opening and closing inventory of a given material. It does not check the rundown of the intra-month inventories. Just because PIMS found a plan in the model feasible, it does not mean it will be feasible in real life (scheduling) as well. A detailed explanation covering the gap between planning and scheduling models in process industries can be found in \cite{harjunkoski2014scope}.

This gap between planning and scheduling can be illustrated by the following example:

We have two units, Unit A and Unit B. Unit A is producing the feedstock of Unit B. We have an inventory tank for this material, with minimum of $5$ kt and maximum of $60$ kt storage capacity. We are going to examine several scenarios in a 30 day long period.

\begin{figure}[h]
\begin{minipage}{0.48\textwidth}
    \includegraphics[width=0.98\textwidth]{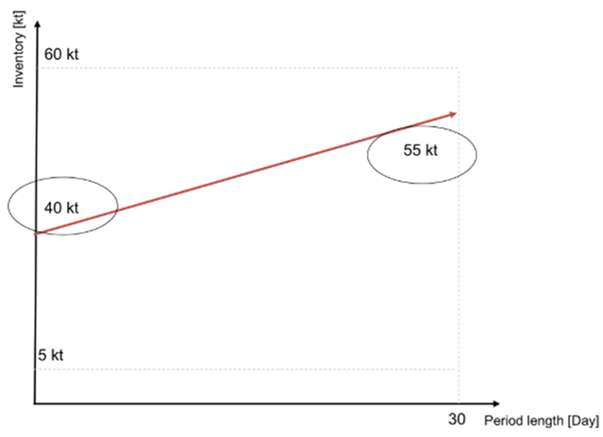}
  \small Base scenario: Inventory change is \\ $+15$ kt
  \label{fig:realchallenges1}
\end{minipage}
\hfill
\begin{minipage}{0.48\textwidth}
    \includegraphics[width=0.98\textwidth]{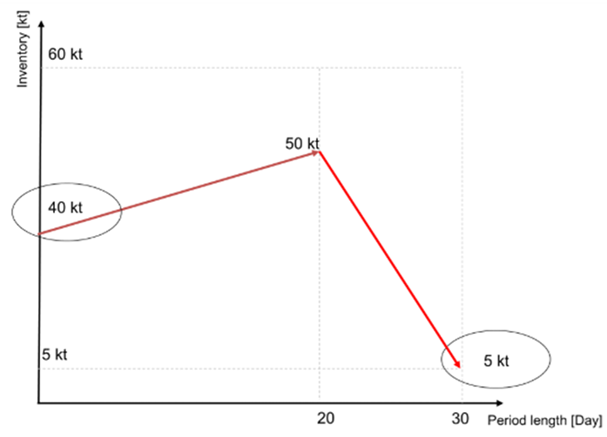}
  \small Scenario A: The shutdown occurs on the 20th of the month.
  \label{fig:realchallenges2}
\end{minipage}%
\vspace{3mm}
\end{figure}

Base scenario: Unit A and B are operating on maximum capacity: Unit A (feed producer) $= 150$ kt/month ($5$ kt/day), Unit B (consumer) $= 135$ kt/month ($4.5$ kt/day), Inventory change $= 15$ kt/month ($0.5$ kt/day).

If we would like to analyze a $10$ day long shut down of Unit A, the monthly balance would look like this from optimization point of view: Unit A $= 100$ kt/month, Unit $B = 115$ kt/month, Inventory change $= -35$ kt/month. 

In the actual scheduling, we can have several different scenarios effecting the feasibility of the plan:

Scenario A: Unit A has a planned shut down for 10 days, from the 20th of the month: Inventory change $= -35$ kt/month: $0.5$ kt/day until the 20th, $-4.5$ kt/day from the 20th. This scenario is still feasible, but reaches the inventory minimum by the end of the period.

\begin{figure}[h]
\begin{minipage}{0.48\textwidth}
    \includegraphics[width=0.98\textwidth]{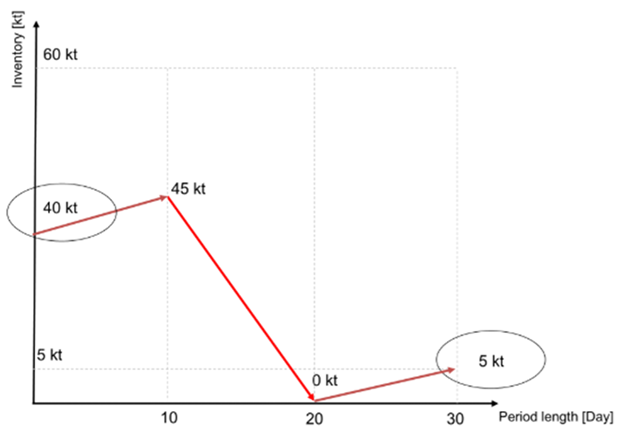}\vspace{2mm}
  \small Scenario B: The shutdown occurs on the 10th of the month (infeasible, because we go below the inventory minimum).
  \label{fig:realchallenges3}
\end{minipage}
\hfill
\begin{minipage}{0.48\textwidth}
    \includegraphics[width=0.98\textwidth]{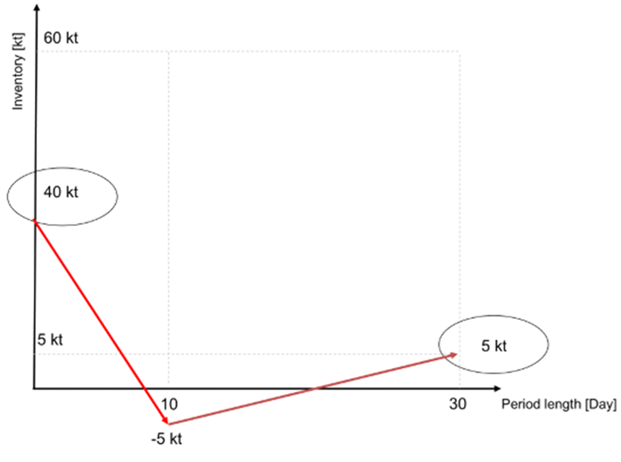}
  \small Scenario C: The shutdown occurs on the start of the month (infeasible, because we go below the inventory minimum).
  \label{fig:realchallenges4}
\end{minipage}%
\vspace{3mm}
\end{figure}

\begin{wrapfigure}{r}{0.5\textwidth}
  \vspace{-18mm}
  \begin{center}
    \includegraphics[width=0.48\textwidth]{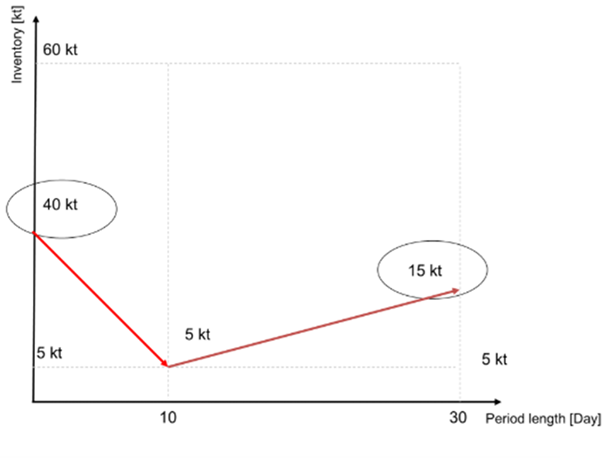}
  \end{center}
  \vspace{-10mm}
  \caption{Optimal Scenario: The shutdown does not cause an inventory deficit or overflow.}
  \label{fig:realchallenges5}
  \vspace{-5mm}
\end{wrapfigure}

Scenario B: Unit A has a planned shut down for 10 days, from the 10th of the month: Inventory change  $= -35$ kt/month $0.5$ kt/day until the 10th. $-4.5$ kt/day between the 10th and the 20th, $0.5$ kt/day from the 20th. The scenario is infeasible, because we go below the inventory minimum.

Scenario C: Unit A has a planned shut down for 10 days, from the 1st of the month: Inventory change = $-35$ kt/month, $-4.5$ kt/day until the 10th, $0.5$ kt/day from the 10th. The scenario is infeasible, because we go below the inventory minimum. To avoid this, we will need to set a $10$ kt lower throughput on Unit B, and a $10$ kt higher closing inventory compared to the Base scenario.

These, and similar settings that are overarching the discrepancy between planning and scheduling have to manually be input into the LP optimization in an iterative manner and require experience and a close cooperation between the corresponding teams.

Interpreting results, understanding connections and the economic drivers require deep domain expertise, analytical thinking, and years of experience. The aim of the Anomaly Detection tool is to aid us in this process, by evaluating past cases and recognizing similar patterns, just like a seasoned expert.

\subsection{Detecting unusual values}

Any value or pattern its own could never be considered an "anomaly" without knowing how this value usually behaves. This is why considering historical data is necessary for Anomaly Detection.

For example, if a price of a product in the current period is $500\$$, but it used to move in the range $100\$-200\$$, then we can be almost sure that it is an anomaly. Though this also depends on how many historical cases we collected. If the $100\$-200\$$ range originates from only a couple datapoints, then the $500\$$ value may not be considered as anomalous as if we had many historical examples. This leads to further considerations in the actual methodology (See Section \ref{section:anomdet}).

\subsection{About the input data}\label{section:inputdata}

In AspenTech PIMS we are differentiating between two types of data in the $\mathbf{A}$ matrix and the $\mathbf{b}$ vector: structural and transactional. 

Structural data describes the complete asset setup from the refinery configuration, technology, to logistics and market points. It is also referred to as “the model” itself. These structures are relatively fix and not changing as often, only when new units are added, existing ones revamped or corrected due to differences compared to reality are found.

Since the plans' structures are constantly changing as units are revamped, this may lead to new variables introduced or removed. Therefore the size of $\mathbf{A}$ and $\mathbf{b}$, as well as $\mathbf{x}$ are subject to change, which leads to further difficulties when applying the anomaly detector on historical data (See the last $3$ paragraphs of Section \ref{section:pairselection}).

Transactional data on the other hand describes the actual operational parameters of the refinery, from available unit capacities, inventory levels, logistical constraints and technological parameters to crude supply, contracted sales and purchases and external environment. These are called “input data” as they are changing from month-to-month and we require data supply to keep them updated. The anomaly detection tool is aimed at these. 

In MOL Group's PIMS model we are investigating the following types of input data with the anomaly detection tool:

\begin{itemize} 
\item Sales – materials sold (from the from refinery) 
\item Purchase – materials purchased (to the refinery) 
\item Capacity – refinery unit capacities  
\item Proclim – refinery process limitations 
\item Bounds – bounded variables to represent various constraints 
\item Transfer – material transfer between plants and depos 
\item Blending – fuel blending numbers 
\item Inventory – refinery and depo inventories 
\item Material Balance – material balance of all streams 
\end{itemize}

Some of the constraints required for these categories are non-linear by nature, and additional sequential linearizations with piecewise linear functions are required.

\subsection{Business opportunities}

Although a comprehensive methodology and several reports were developed in MOL Group’s optimization team to support the understanding of the drivers and opportunities of this extremely complex optimization dilemma, these face two challenges:

\begin{enumerate}
\item Time-series: the previous methods were focusing on single case data – there was no connection provided to historical results, therefore any pattern or anomaly recognition could only be performed by the optimizer expert, and as such was limited by the expert's time, experience and analysis skills. 
\item Limited number of KPIs monitored: although the most important results were checked and monitored in simple dashboards, many relations remained hidden from the tens of thousands of data points.
\end{enumerate}

\subsection{Technical challenges}

Any unsupervised machine learning algorithm, just like Anomaly Detection, expects clean data, with all input values clearly available for all historical cases. However, the data listed in Section \ref{section:inputdata} was available though multiple different SQL data tables, which not only needed to be collected, but also joined through multiple types of key columns.

This procedure was encoded, with an additional interface for modelers to change which SQL tables data is extracted from, and on which keys they are joined.

\section{Anomaly Detection methodology}\label{section:anomdet}

Multiple Machine learning algorithms have been developed for anomaly detection, grouped into supervised and unsupervised learning. See a detailed overview of process fault detection and diagnosis in chemical engineering in \cite{venkatasubramanian2003review}. We are now interested in the unsupervised anomaly detection case, since we do not have labeled data. DBSCAN and the Clustering-based Local Outlier Factor are two clustering-based techniques for outlier detection. Neighbor-based techniques consider outliers as lonely observations "far" from others observations. K-nearest neighbors (KNN) is a simple technique in which the average distance to $k$ nearest neighbors is computed based on distance metrics such as Euclidean, Manhattan or other distances. The points with the biggest average distances are marked as outliers \cite{dang2015distance}. Additionally, dimensionality reduction methods, such as PCA; as well as statistical monitoring methods are suggested in \cite{qin2012survey}.

Density-based algorithms cluster the points based on the distance between a data point and a cluster. Singular points, or points in the lower density regions, are marked as outliers. Local Outlier Factor and Connectivity-Based Outlier Factor are two widely used density-based algorithms that are very efficient at finding high-density regions (normal instances) and low-density regions (outliers). Finally, angle-based methods consider not only the distance between points but also the directions of the distance vectors. An angle-based outlier factor is calculated, and data with small values are labeled as anomalies. \cite{kriegel2008angle}.
Proximity measures are easier to calculate than to fit statistical distributions, but the high dimension makes the distance-based methods unusable and also the distribution fitting hard. This is why in the next subsection we tackle the high-dimensional problem by using the most informative pairs.

\subsection{Univariate Anomaly Detector - ECOD}\label{section:ECOD}

The ECOD Anomaly Detector was originally developed by Z. Li et al. in 2022 \cite{li2022ecod}. The idea was to calculate the (one-dimensional) cumulative distribution functions (c.d.f.’s) of each input variable in the historical (train) data $F_i$ and compare each input variable of the new monthly plan's data to these $F_i$'s. If the value of the c.d.f. $F_i(x)$ at the new datapoint $x$ is extraordinarily high or extraordinarily low, then it must be an anomaly.

\begin{figure}[h]
    \centering
    \includegraphics[width=0.95\linewidth]{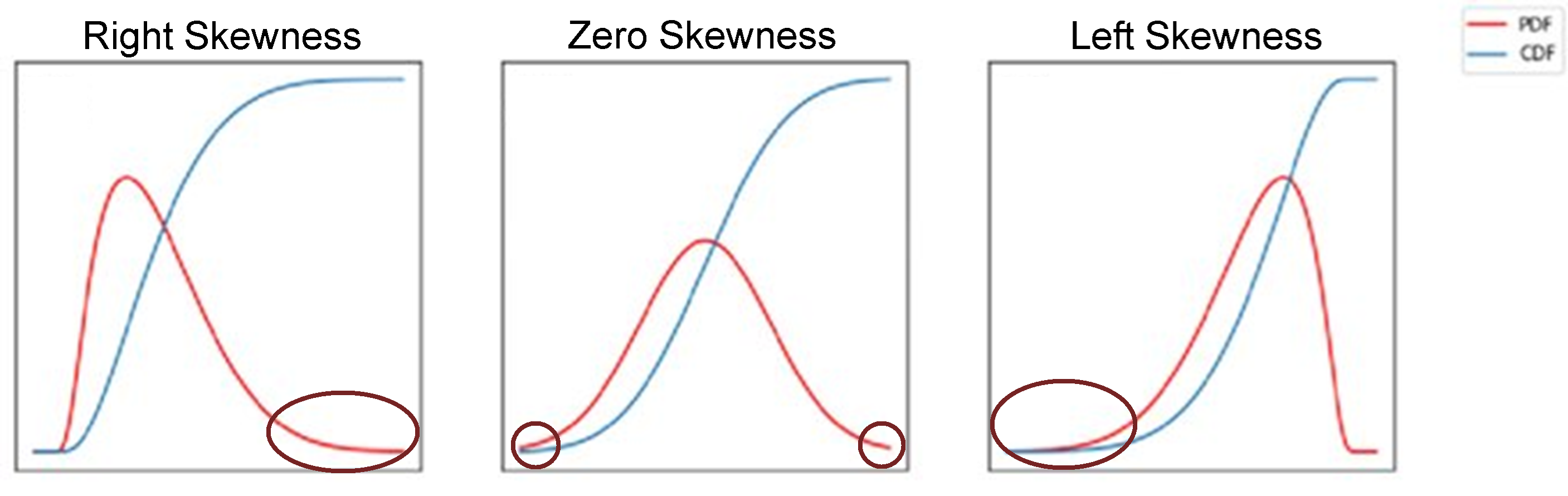}
    \caption{The location of anomalies expected, depending on the skewness of the historical data's CDF. Skewness captures which side the long tail is located on, which is where anomalistic values could show up (dark red).}
    \label{fig:skewness_anomalies}
\end{figure}

If the difference between $0$ and $F_i(x)$ (on the left side of the c.d.f. function), or the difference between $1$ and $F_i(x)$ (on the right side of the c.d.f. function) is extremely small, then the given value is an anomaly. In fact, this difference could be used as an “Anomaly Score”. This is useful because sorting by this Anomaly Score allows us to find the largest anomalies. The smaller this value, the farther away the new input value is from its old distribution, so the more likely that it is an anomaly. Moreover, in the multivariate case, these scores are comparable to each other, which makes it possible to select the most anomalous input indices.

More precisely, \cite{li2022ecod} suggests that we find the skewness of $F_i$. In the case of right skewness, the anomalies accumulate on the left of $F_i$ while in the case of left skewness, the anomalies will be on the right. Therefore, depending on the skewness of $F_i$ the following value could be used as an anomaly score:

$$\begin{cases}
F_i(x) \quad\quad\ \ \, \text{in the case of right skewness} \\
1-F_i(x) \quad \text{in the case of left skewness} \\
\end{cases}$$

However, in case of anomalies, the listed $F_i(x)$ and $1-F_i(x)$ values get exponentially small in the tails of the c.d.f., so the use of $-\log_2$ was suggested to yield the final ECOD anomaly score of 

$$\text{ECOD}(x)=\begin{cases}
-\log_2(F_i(x)) \quad\quad\ \ \, \text{in the case of right skewness} \\
-\log_2(1-F_i(x)) \quad \text{in the case of left skewness} \\
\end{cases}$$

This way, large ECOD Anomaly Scores mean large anomalies, small scores mean regular values (non-anomalies). 

This is as far as the original ECOD methodology goes, however, when actually applying this method on real data, we have found the following shortcomings: 

\begin{enumerate}
    \item Very often, the value of $\log(0)=-\infty$ would be encountered. That means that (in the case of right skewness), $F_i(x)=0$, meaning that in the historical data, every value of the given input was greater than $x$ or in other words, we have never seen such a small $x$ value in the past. Or in the case of left skewness, we have never seen such a large $x$ value in the past. So for example, when the entire system is flawed, e.g. it encounters a loophole of being able to transfer/buy/sell infinitely, such large values can occur very frequently. All of these cases resulted in many ECOD Anomaly Scores of $-\infty$, which could no longer be compared to each other.
    \item Quite often, the $F_i$ functions were calculated from only a few data points, which meant that the ECOD  Anomaly Scores were not telling the whole truth. This has also yielded cases of $\log(0)=-\infty$ with the only difference that sometimes entirely correct values would also gain an infinitely large anomaly score.
\end{enumerate}

To solve these issues, we have applied the following modifications:

\begin{enumerate}
    \item We have redefined the ECOD Anomaly Score to be

    \begin{equation}\label{eq:ecodprime}
     \text{ECOD}'(x):=\begin{cases} \frac{c \cdot |x-\hat{x}|}{s(x)^k} \ \qquad\quad \text{if ECOD}(x) \text{ is undefined} \\
    \text{ECOD}(x) \quad\ \  \text{otherwise}
    \end{cases}
    \end{equation}

    Here, $c$ is a scaling factor that controls how important it is for the anomaly score to indicate that the given value does not exist in the train data. By default, it is set to $c=10$, but I have also made it possible for the user to change it in the future. Moreover, if the first case is chosen, the output also indicates that this $x$ value was never seen in the input dataset, which I have called “AA type anomalies”. A cutoff was chosen for the rest of the cases. The first $5000$ anomalies found this way (sorted by the ECOD Anomaly Score) were called “A type anomalies”. 

    The $k$ exponent of $s(x)$ is also a parameter. By default, one could argue that it should be set to $1$, to standardize $|x-\hat{x}|$. However, we have found that there are various data types whose higher moments are significant, that is, their variance also varies a lot. Therefore, in reality, after some experiments, we have chosen a number greater than $1$, namely $k=1.4$. This calms the variance and makes the $\text{ECOD}'$ score between different types of data slightly more comparable.

    \item We have introduced three more cutoffs to the system. Firstly, constant c.d.f.s were discarded. Secondly, the train data would only contain c.d.f.s that are calculated from at least $k$ historical samples (where we have chosen $k=5$ for this parameter -- any less than this, and fitting a linear regression to this number of points becomes too ambiguous). Thirdly, the proportion between the datapoints that the given c.d.f. is calculated from and the number of monthly plans the historical data contains must be at least $p$ (where we have chosen $p=0.05$ for this parameter -- any less than this, and the variable is no longer represented in enough monthly plans). These two empirical cutoffs avoid calculating the c.d.f. from too few data points, and we will not be checking for anomalies in these cases.
\end{enumerate}

\subsection{Multivariate Anomaly Detection}
Anomaly Detection in high dimensions is a multifaceted problem, and the choice of detection techniques is strongly influenced by how anomalies are defined, as well as by the nature of the input data and the expected outputs. These factors lead to a wide range of problem formulations, each requiring different analytical and computational approaches. The presence of anomalies, which may be data flaws or measurement errors, can lead to model misspecification and misleading results in the case of classical models such as Linear Programming. In contrast, there are cases when anomalies are the main carriers of important information, which may lead to significant economic opportunities. 

In the following subsections, our anomaly detection methods are introduced that help the experts improve the quality of data and enhance the work of the decision-makers. 

\subsubsection{A pair-selection method for bivariate Anomaly Detection}\label{section:pairselection}

In unsupervised Anomaly Detection in high dimensions, a main problem is caused by the sparsity of the data. To overcome this, we apply the idea introduced in \cite{horvath2020copula} to use an informative subset of the bivariate marginal distributions. At the core of this method is the goal of finding an approximating multivariate probability distribution of the form
\begin{equation}
\label{treedistr}
p(x_{1},\ldots,x_{n})=\frac{\underset{(i,j)\in T}{\prod }p(x_{i},x_{j})}{%
\underset{i\in V}{\prod }\left[ p(x_{i})\right] ^{v-1}}
\end{equation}
defined by only bivariate probability distributions, where $v_i$ denotes the degree of vertex $i$ in a spanningtree $T$. These bi-variate marginals can be approximated from the data. The main idea is to find that distribution of form \ref{treedistr} which minimizes the Kullback-Leibler divergence to the real data. In  \cite {chow1968approximating} for discrete probability distributions it is shown  that finding the maximum weight spanning tree in a complete graph, in which the vertices are associated with the random variables and the weights are defined by the mutual information of connected variables, determines the bivariate marginals included in the approximation. Now we do this by using Kendall tau applied to the data. The Anomaly Detection method presented in \cite{horvath2020copula} employs a parametric approach for modeling the bivariate marginals by separately fitting the bivariate copulas and the marginals.

Since in our case we deal with an extremely large amount of data and variables, we decided to use Gaussian bivariate marginals, which are faster to fit and are more robust.

In this paper, we introduce two techniques to identify anomalies in the bivariate marginals, as described in the following subsections.

In the Section \ref{section:ECOD}, we have seen how one-dimensional c.d.f.s $F_i$ can be used to find rare, unusual values, which we have called anomalies. In the following two subsections, we will be using certain two-dimensional marginal random vectors $(X_i,X_j)$, where $i$ and $j$ are two different indices of the column vectors of the LP model.
Now due to the high-dimensionality of the data we will explain how we select the important pairs $(X_i,X_j)$; then explain how to apply two different types of anomaly detectors.

The used methodology goes the following way:

\begin{enumerate}
    \item Pre-define groups of input pairs we would like to calculate anomalies for. In the case of large oil refinery models, unfortunately, calculating an Anomaly Score for all possible pairs seems to not be feasible, not to mention the interpretation of a high Anomaly Score between certain variables becomes fuzzy. For example, there is no point in comparing the Marginal Value of a purchase to the capacity of a completely different plant – even if such a pair has unusual values, the planning team wouldn’t be able to use this information. Therefore, a manually defined pre-selection of variable groups is necessary, which in our case, reduced the number of pairs to look through from $\sim 7$ billion to $\sim 4$ million. This was done by the expert modelers, then afterwards the Anomaly Detector automatically does this filtering every time it is run.
    \item Further filter the potential pairs using Kendall’s Tau. Calculate a Kendall correlation matrix between all pairs defined in the first step. To do this, we create a complete graph with nodes for each variable, and edge weights equal to the absolute value of Kendall’s Tau between each connected pair. Then we calculate a maximum weight spanning tree, and only keep edges (pairs) that are inside this spanning tree. This avoids listing the same anomaly source twice, and only lists the most significant anomalies. For example, if the values of all of $i$, $j$ and $k$ are extreme, then all pairs $(i,j)$, $(j,k)$, and $(j,i)$ would be listed as anomalies. Using a spanning tree however, one of these pairs – the one with the lowest two-dimensional anomaly score – will be discarded. Finally, we obtained pairs with mostly high Kendall correlation. After this procedure, some low-Kendall correlation pairs have still remained, so we decided to implement a cutoff, $K$, and remove pairs that have an absolute Kendall correlation below $K$ (in our case, $K=0.4$ -- the correlation coefficient indicates no significant correlation below this value). This makes sure we are really only dealing with existing relationships. In our case, this process reduced the number of potential pairs to look through from $\sim 4$ million to $\sim 20$ thousand.
    \item Then, we are preparing for the two multidimensional Anomaly Detection methods described in Sections \ref{section:linreg} and \ref{section:mvs}. For all of the $(x,y)$ pairs obtained, calculate the best fitting line $y=ax+b$, an $R^2$ linear fit score of this line, an average $\overline{e}$ and a variances $s_e^2$ over a fit line (see Formula \ref{formula:linreg_saved}). Then, for all the pairs obtained, also calculate $\mu$, the 2-dimensional historical mean of the pair's values, and a $2 \times 2$ covariance matrix $V$ from the same historical data.
    
\end{enumerate}
    
In our experience, calculating such a large correlation matrix in Step 2 takes several hours. To avoid having to wait this much every time the Anomaly Detector runs, we have applied the train – retrain methodology. This means that to “fit” this Anomaly Detector, we first need to train (or retrain) it, where Steps 1-3 will be calculated. When applied to new data (tested), only the steps described in Sections \ref{section:linreg} and \ref{section:mvs} will be applied.

There is however one caveat in Step 2 of this methodology. In reality, there are a huge number of missing values. For one, not all input values are present in all monthly plans, in fact, every plan is usually missing a fair number of input values. However, a correlation matrix calculation expects input vectors of the same length, and we are dealing with non-equal length input vectors (the amount of missing values is different between each input). Matching up the correct values of each input variable's historical range for every element of the correlation matrix turned out to be an incredibly demanding task from a computational point of view.

To resolve this, we are using a penalty function method. This method calculates the correlation matrix normally: It fills in actual missing values with the average of the given input’s values. Then it applies a multiplier to each row and each column of the correlation matrix. This multiplier ranges between $0$ and $1$, and is closer to 1 the more missing values we have. It makes sense to linearly penalize the correlation matrix, namely if for example $30\%$ of the values were missing, then we multiply the given correlation by $1-0.3=0.7$. However, each row and column may need to be multiplied, so we instead multiply by the root of this, so that two roots will cancel out to obtain a linear penalization.

More specifically, we multiply each row and each column by $\sqrt{\frac{n-s}{n}}$, where $s$  is the number of values considered to be “missing”, and $n$ is the number of historical cases. This is the actual correlation matrix we use in Step 2.

This reduced the calculation time of the correlation matrix from more than $10$ hours to around $2-3$ hours.

\subsubsection{Linear Regression Anomaly Detector}\label{section:linreg}

\begin{wrapfigure}{r}{0.4\textwidth}
\vspace{-8mm}
  \begin{center}
    \includegraphics[width=0.38\textwidth]{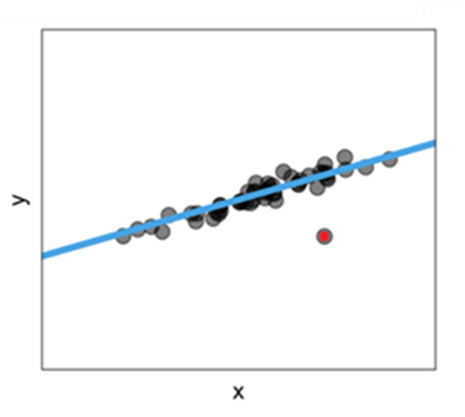}
  \end{center}
  \vspace{-7mm}
  \caption{Example of an 2-dimensional anomaly that would not be detected in one dimension}
  \label{fig:linear_anomaly}
  \vspace{-5mm}
\end{wrapfigure}

Anomalies do not just come alone. Sometimes two different LP model values can individually be in the expected range, but together still be far from the usual (See Figure \ref{fig:linear_anomaly}).

The two coordinates of the red dot represent the current values of the $x$ and the $y$ variables, respectively. When projected to the $x$ axis, the red dot will be lost in the range of all the other black dots, and same for the $y$ direction. Yet the rest of the black dots all lie near the blue line, while the red dot lies far away. In this case, the $y$ value was historically approximately some linear function of $x$, let’s say $y \approx ax+b$, except in the red case.

\begin{figure}[h!]
  \begin{center}
    \includegraphics[width=0.78\textwidth]{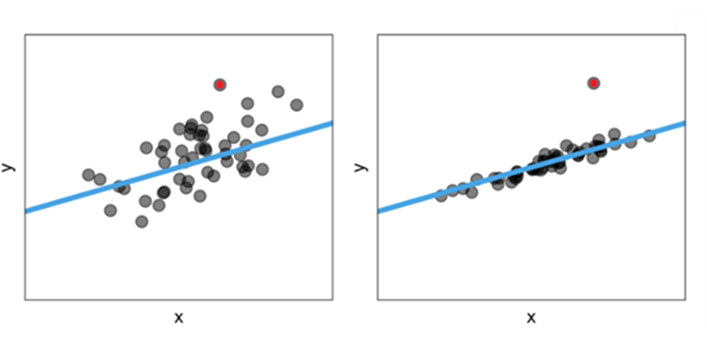}
  \end{center}
  \caption{Each bivariate margin of the Anomaly Score must depend on the variance of the train data. While the red dot is the same distance away from the blue line in both cases, it is a relatively realistic value on the left, and an anomaly on the right.}
  \label{fig:linear_score_comp}
\end{figure}

The goal of the Linear Regression Anomaly Detector is first to find such variable pairs for which historical data would suggest a strong linear correlation (a.k.a., the $y$ variable is usually very close to $ax+b$ for some values of $a$ and $b$), and for these pairs, give the best estimation for $a$ and $b$, and a linear fit score. The linear fit score defines how strong this correlation is – see Figure \ref{fig:linear_score_comp}: In the left case, $x$ and $y$ have a weak correlation, so low linear fit score, and in the right case, this correlation is high. Even though the red dot is the same distance away from the blue line in both cases, since the linear fit score on the left is smaller, the red dot is less of an anomaly than on the right. The linear fit score mentioned is actually the Pearson-correlation ($R^2$) score.
    
When applying the Linear Regression Anomaly Detector, we can read the new input data, run through all $20$ thousand pairs filtered in Section \ref{section:pairselection}, and see if any of them are far outside of the expected $y=ax+b$ line. To quantify this, we calculate a pairwise Anomaly Score, in the following way: First we draw a normal distribution over the $y=ax+b$ line with variance equal to the train data’s variance around this line. Then, the Anomaly Score will be the value taken on by this normal distribution at the new pair’s coordinates. This means that a lower Anomaly Score actually describes a greater anomaly. In terms of formulas, the Anomaly Score of a new pair of values $(x_{n+1},y_{n+1})$  will be following:
$$\text{AnomalyScore}(x_{n+1},y_{n+1})=\frac{1}{\sqrt{2\pi s_e^2}} \exp\left(-\frac{(y_{n+1}-(ax_{n+1}+b)-\overline{e})^2}{2s_e^2}\right)$$
where
    
\begin{equation}
\begin{split}\label{formula:linreg_saved}
\overline{e} &= \frac{1}{n} \sum_{i=1}^n (y_i -(ax_i +b)) \\
    s_e^2 &= \frac{1}{n-1}\sum_{i=1}^n (y_i-(ax_i+b)-\overline{e})^2 \\
    y &= ax+b \text{ is the best fitting line on the } (x,y) \text{ data}
\end{split}
\end{equation}
The parameters $a$, $b$, $\overline{e}$ and $s_e^2$ were calculated and saved in Step 3 of Section \ref{section:pairselection}.

\subsubsection{Multivariate Sampling Anomaly Detector}\label{section:mvs}

In the Multivariate Sampling methodology, we defined scores based on ellipses, obtained from the level sets of a 2-dimensional multivariate distribution, calculated from historical data. On Figure \ref{fig:mvs_cutoff}, the red ellipse is the best fitting ellipse to the data, such that it has the lowest area that still surrounds $99\%$ of the historical data, and only exactly $1\%$ of the historical data is outside. A 2-dimensional normal distribution is defined by a mean vector $\mu$ ($2$ values, the individual averages of the $x$ and $y$ directions), and a $2\times 2$ covariance matrix $V$. These were already calculated in Step 3 of Section \ref{section:pairselection}. Afterwards, the Multivariate Sampling algorithm will be the following:

\begin{wrapfigure}{r}{0.4\textwidth}
  \vspace{-5mm}
  \begin{center}
    \includegraphics[width=0.38\textwidth]{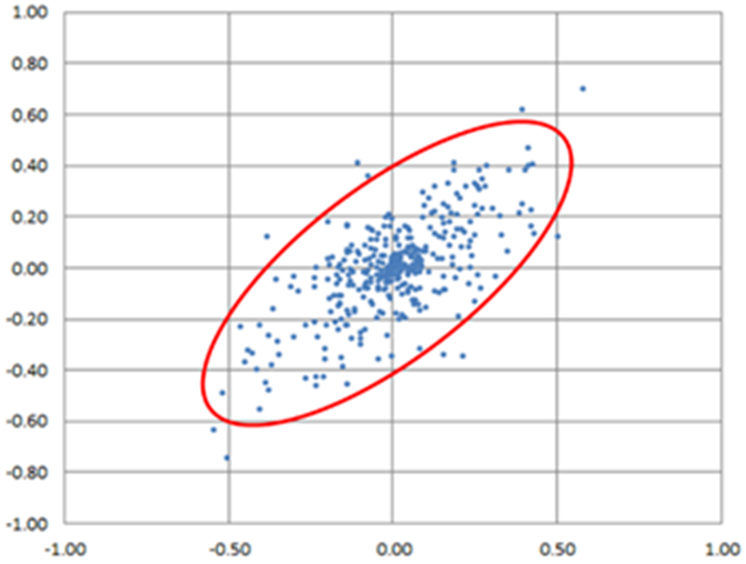}
  \end{center}
  \vspace{-2mm}
  \caption{The 99\% level set of a 2D multivariate normal distribution fitted to a variable pair of historical values.}
  \label{fig:mvs_cutoff}
  \vspace{-2mm}
\end{wrapfigure}

We calculate a pairwise Anomaly Score. A fitted normal distribution already exists, so the Anomaly Score will simply be the value this function takes on at the new data’s coordinates. A $p$-cutoff-value can also be calculated. This value expresses how rare such a pair is. If the $p$-cutoff-value is $0.01$ or $1\%$, then the pair of values is outside the above shown ellipse. If the $p$-cutoff-value is $0.05$ or $5\%$, then the pair is outside the ellipse defined by the $5\%$ cutoff. These two cutoffs are calculated by Multivariate Sampling (the namesake of the method). Since the data is sparse, we sample $1500$ normally distributed two-dimensional values using a Monte-Carlo simulation, with the same mean $\mu$ and covariance $V$ as the historical data. (This $1500$ can be changed according to the size of the original problem.) In terms of formulas, the Anomaly Score of a new pair of values $(x_{n+1},y_{n+1})$ will be following:

\vspace{-3mm}
\begin{align}\label{formula:mvs_saved}
\begin{split}
&\text{AnomalyScore}(x_{n+1},y_{n+1}) = \\ &= \frac{1}{\sqrt{(2\pi)^2 \det V}} \exp\left(-\frac{1}{2}\left((x_{n+1},y_{n+1})-\mu\right)^T V^{-1} \left((x_{n+1},y_{n+1})-\mu\right)\right)
\end{split}
\end{align}
where $\mu$ and $V$ are calculated and saved in Step 3 of Section \ref{section:pairselection}.

\subsection{Assembling the full Anomaly Detector}

\begin{figure}[h!]
\captionsetup{singlelinecheck=off}
    \centering
    \includegraphics[width=0.95\linewidth]{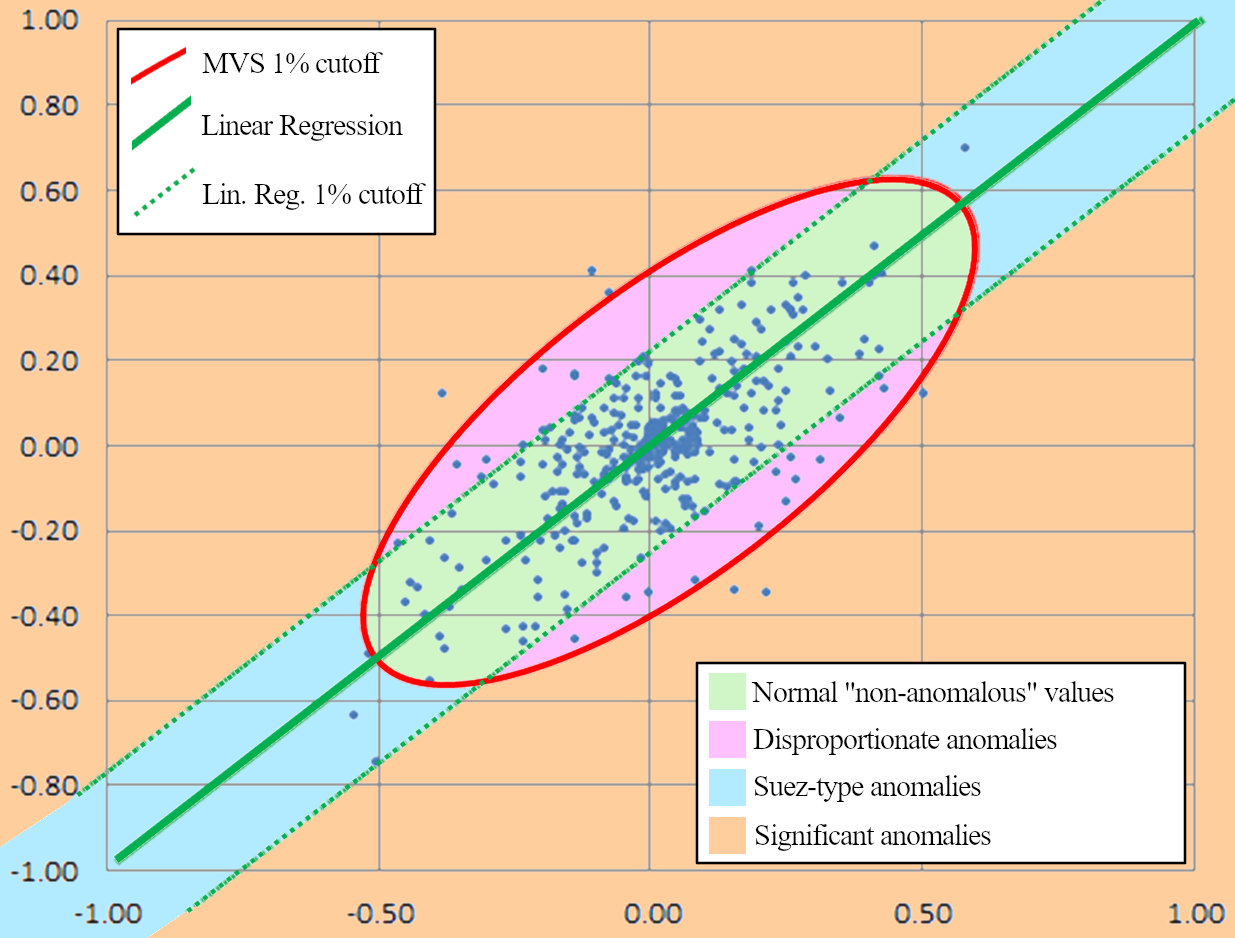}
    \caption[figure]{The regions determined by the Anomaly Detector for each pair of historical values considered.
    
    \begin{itemize}
    \item The red ellipse is determined by Multivariate Sampling, and it is the best fitting ellipse to the datapoints such that $1\%$ of the historical data is outside.
    \item The green line is determined by Linear Regression and it is the best fitting line to the data. The dashed green line is chosen such that $1\%$ of the historical data is outside of it.
    \item The central green region is the "non-anomalous" region.
    \item The two purple regions are where Disproportionate Anomalies are found.
    \item The two blue regions are where Suez-type Anomalies are found.
    \item The two large orange regions are where Significant Anomalies are found.
    \end{itemize}
    }
    
    \label{fig:anomaly_regions}
\end{figure}

Figure \ref{fig:anomaly_regions} contains an example historical range of a pair of inputs considered. We then apply two Anomaly Detection methods introduced in Section \ref{section:linreg} and Section \ref{section:mvs}. Both of these methods have a $1\%$ cutoff based on historical data. (Outside of this cutoff, historically, only $1\%$ of the data was found.) The intersections of these regions are interpreted as different types of anomalies.

\begin{itemize}
    \item \textbf{Non-anomalous region (green)}: The region that neither method considers an anomaly.
    \item \textbf{Significant anomalies (orange)}: The region that both methods consider an anomaly. Values found here are very likely to be anomalous, therefore the "significant" marker.
    \item \textbf{Disproportionate anomalies (purple)}: The region that Linear Regression considers an anomaly, but Multivariate sampling does not. Since value pairs "close" to the Linear Regression line have approximately the same proportion, values "far away" from this line can be considered disproportionate.
    \item \textbf{Suez-type anomalies (purple)}: The region that Linear Regression does not consider an anomaly, but Multivariate Sampling does. In these regions, value pairs are both either unexpectedly large or unexpectedly small (e.g. two prices skyrocketed), but compared to each other, they are still about the same. The name comes from the fact that in 2021, famously, a ship got stuck in the Suez-canal, which lead to such anomalies - prices went up, but all by approximately the same amount. \cite{suez}
\end{itemize}

\begin{landscape}

\begin{figure}
    \centering
    \includegraphics[width=0.99\linewidth]{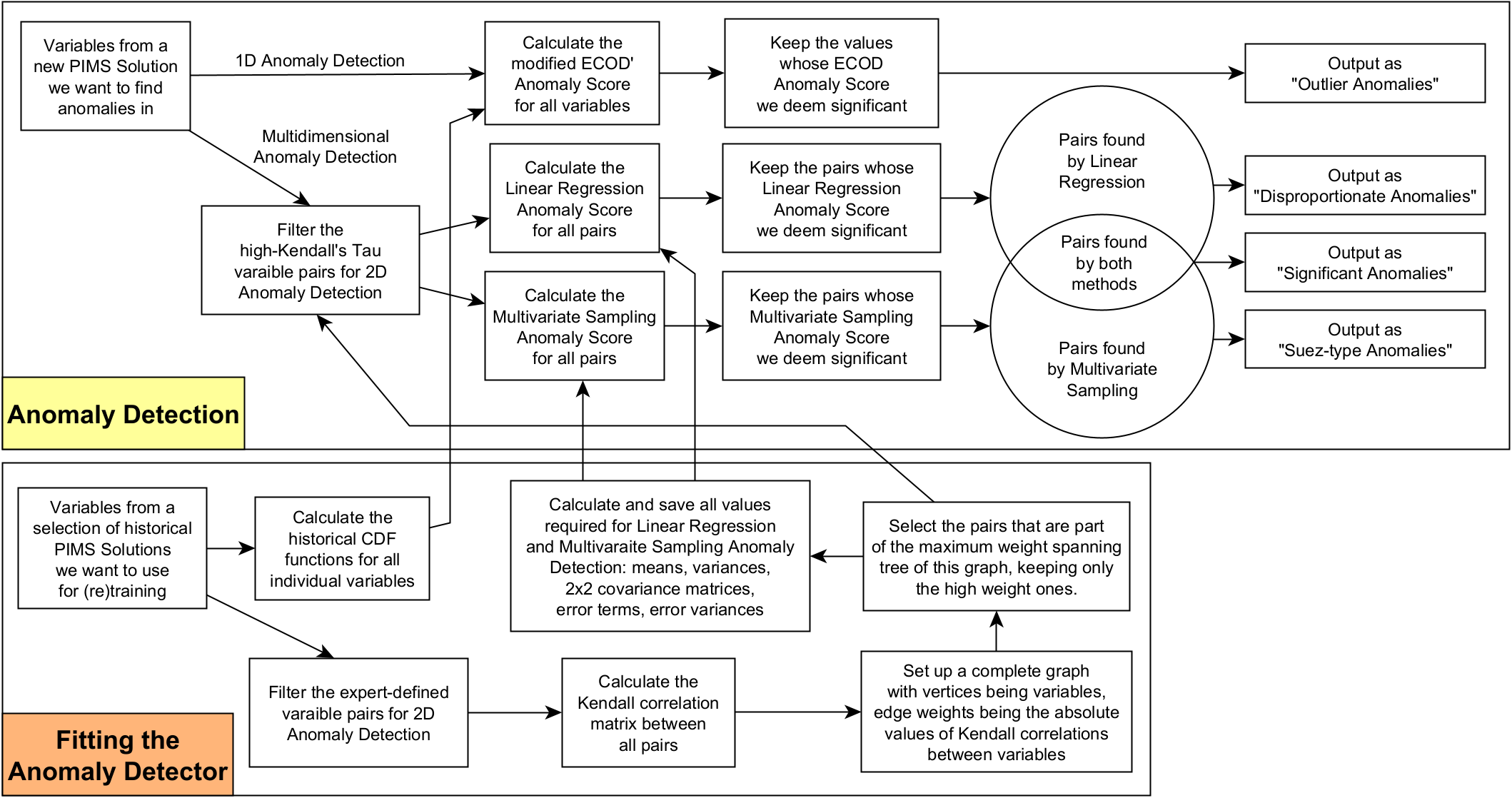}
    \caption{The outline of the Anomaly Detection process. First the Anomaly Detector is fit using historical PIMS Solutions. This calculates the 1D CDFs, important \& highly-correlating pairs to look through in the multidimensional case, and the parameters $a,b,\overline{e},s_e^2,\mu,V$ for each considered pair $(x,y)$. Then the Anomaly Detector uses these pre-calculated values to find the one- and two dimensional Anomaly Scores for each considered value or pair. The resulting 2D anomalies are transformed via the Venn diagram shown for an easier interpretation of the Anomaly types.}
    \label{fig:AnomDet_methodology}
\end{figure}

\end{landscape}

As mentioned in Section \ref{section:pairselection}, the program needs to be periodically retrained. We chose a period of $3$ months for this. Figure \ref{fig:AnomDet_methodology} shows what the (re)training algorithm does, and how it affects the main Anomaly Detection algorithm.

\section{Algorithms comprising the Anomaly Detection methodology}

In this section, we will describe the Anomaly Detection train and test algorithms in full detail and calculate their algorithmic complexities. As a reminder, Anomaly Detection is part of unsupervised learning. Here, "testing" simply refers to applying the Anomaly Detector on a new case.

\subsection{The Anomaly Detection train algorithms}

Throughout this subsection, we will denote the $i$'th column of a dataset $D$ by $D[i]$.

The input of all train algorithms is the collection of historical data, which will be denoted by $D\in\mathbb{R}^{m\times n}$ here. This dataset has the following indices:

\begin{itemize}
\item Row indices for historical data $i \in \{1,\dots,m\}$
\item Rolumn indices for Tags (dtype, opttype, $j$), where
\begin{itemize}
\item dtype $\in \{$Sale, Purchase, Capacity, Proclim, Bound, Transfer, Blending, Inventory, Material Balance$\}$
\item opttype $\in \{$Activity, MarginalValue, MinValue, MaxValue, Opening, Closing$\}$
\item $j \in \mathbb{N}^+$
\end{itemize}
\end{itemize}

For all MaxValue opttype cells in $D$, if the cell's value is empty or infinity, replace it with $\max D + 1$.

\vspace{3mm}
\begin{breakablealgorithm}
\caption{ECOD Anomaly Detector training}\label{alg:ecod_train}
\textbf{Input:} The $D$ dataset.
\begin{algorithmic}[1]
\State Let $S = \{c_1,\dots,c_n\}$ be a new set of columns obtained from the columns of $D$ by omitting all NaN values.

\For{$j \in \{1,\dots,n\}$}
    \State Let $F_j$ be the empirical cumulative distribution function of $c_j$.
    \State Let $\mu_j$ be the empirical mean of $c_j$.
    \State Let $\sigma_j$ be the empirical standard deviation of $c_j$.
    \State Let $s_j$ be the empirical skewness of $c_j$.
\EndFor

\end{algorithmic}
\textbf{Outputs:} $F_1,\dots,F_n$, $\mu_1,\dots,\mu_n$, $\sigma_1,\dots,\sigma_n$, $s_1,\dots,s_n$.
\end{breakablealgorithm}

\vspace{3mm}
\begin{breakablealgorithm}
\caption{Finding the most informative pairs}\label{alg:pairs}
\textbf{Inputs:} The $D$ dataset, $\mathcal{P}$ the set of expert-defined ((dtype$_1$,opttype$_1$), (dtype$_2$,opttype$_2$)) pairs to search anomalies in, Relative standard deviation cutoff (by default, $0.01$), Kendall-correlation cutoff (by default, $0.4$).
\begin{algorithmic}[1]
\State Let $\mathcal{M}$ be an empty list where the output will be collected.
\For{$(L,R) \in \mathcal{P}$}
    \State Let $D_L \in \mathbb{R}^{m \times n_L}$ be the dataset of columns of $D$ with (dtype,opttype)$=L$.
    \State Let $D_R \in \mathbb{R}^{m \times n_R}$ be the dataset of columns of $D$ with (dtype,opttype)$=R$.
    \State $\forall i \in \{1,\dots,n_L\}$, let $m_L(i)$ be the number of missing values in $D_L[i]$.
    \State $\forall j \in \{1,\dots,n_R\}$, let $m_R(j)$ be the number of missing values in $D_R[j]$.
    \State $\forall i \in \{1,\dots,n_L\}$ let $p_L(i):=\sqrt{1-m_L(i)/m}$
    \State $\forall j \in \{1,\dots,n_R\}$ let $p_R(j):=\sqrt{1-m_R(j)/m}$
    \State For each column of $D_L$ and $D_R$, fill out the missing values of the column with the mean value of the column.
    \State Remove all columns of $D_L$ and $D_R$ whose relative standard deviation is is below the Relative standard deviation cutoff.
    \For{$i \in \{1,\dots,n_L\},\ j \in \{1,\dots,n_R\}$}
        \State Let $C_{LR}(i,j) :=$ The Kendall-correlation between $D_L[i]$ and $D_R[j]$.
        \State Let $\tilde{C}_{LR}(i,j) :=C_{LR}(i,j)p_L(i)p_R(j)$ the penalized corr. matrix.
        \State Let $A_{LR}(i,j) :=|\tilde{C}_{LR}(i,j)|$ the absolute penalized corr. matrix.
        \State If $A_{LR}(i,j) = \pm \infty \rightarrow A_{LR}(i,j) := 0$
    \EndFor
    \State Let $G=(I_L\cup I_R,\tilde{E})$ be a bipartite graph with vertices $V=I_L\cup I_R$, where $I_L$ are the column indices of $D_L$, $I_R$ are the column indices of $D_R$, and edges $e_{ij} \in \tilde{E}$ connecting every element of $I_L$ to every element of $I_R$ with edge weights $A_{LR}(i,j)$.
    \State Let $T=(V,E)$ be a maximum spanning tree of $G$.
    \State Discard all edges $e \in E$ of $T$ with edge weight less than the Kendall-correlation cutoff.
    \For{$e=(i,j) \in E$}
    \State Add the following pair to the $\mathcal{M}$ list: $((d_1,o_1,i),(d_2,o_2,j))$, where $L=(d_1,o_1), R=(d_2,o_2)$.
    \EndFor
\EndFor
\end{algorithmic}
\textbf{Output:} The $\mathcal{M}$ list of most informative pairs.
\end{breakablealgorithm}

\vspace{3mm}
\begin{breakablealgorithm}
\caption{Linear Regression and Multivariate Sampling Anomaly Detector training}\label{alg:linreg_mvs_train}
\textbf{Inputs:} The $D$ dataset, $\mathcal{P}$ be the set of expert-defined ((dtype$_1$,opttype$_1$), (dtype$_2$,opttype$_2$)) pairs to search anomalies in, Minimum sample amount cutoff (a number, e.g. $5$, or a proportion of all $m$ samples, e.g. $1\%$) Relative standard deviation cutoff (by default, $0.01$), $v$ Multivariate sampling size (by default, $1500$), $\omega$ p-Value cutoff (by default, $0.99$)
\begin{algorithmic}[1]
\State Let $\mathcal{M}:=$ The output of Algorithm $\ref{alg:pairs}$, the most informative pairs to search anomalies in.
\State Let $\mathcal{L}$ an empty list where the output will be collected.
\For{$(L,R) \in \mathcal{M}$}
    \State Let $l$ be the column of $D$ with column index $L$.
    \State Let $r$ be the column of $D$ with column index $R$.
    \State Restrict the $l$ and $r$ vectors to indices where both are not NaN. Let us denote the resulting length of $l$ and $r$ by $m_2$.
    \If{$m_2 \ge $ Minimum sample amount cutoff and the relative standard deviation of both $l$ and $r$ are both greater than the Relative standard deviation cutoff}
        \State Let $\xi := [l\ \ r]^T$
        \State Let $\mu_l :=$ the mean of $l$, $\mu_r:=$ the mean of $r$, and $\mu = [\mu_l \ \ \mu_r]^T$.
        \State Let $V :=$ the covariance matrix of $(l,r)$.
        \For{$k \in \{1,\dots,m_2\}$}
            \State $p_k := \frac{1}{\sqrt{2\pi}\det(V)}\exp(-\frac{1}{2}(\xi[k]-\mu)^T V^{-1}(\xi[k]-\mu))$
        \EndFor
        \State Let $\eta_1,\dots,\eta_v$ be $v$ independent samples of the $N(\mu,V)$ distribution.
        \For{$k \in \{1,\dots,v\}$}
            \State $q_k := \frac{1}{\sqrt{2\pi}\det(V)}\exp(-\frac{1}{2}(\eta[k]-\mu)^T V^{-1}(\eta[k]-\mu))$
        \EndFor
        \State Let us sort $p_1,\dots,p_{m_2}$ in a descending order to obtain $p_{\phi(1)}, \dots, p_{\phi(m_2)}$.
        \State Let $\hat{X}:=(\xi_{\phi(1)},\dots,\xi_{\phi(m)})$.
        \State Let us cut off the rightmost $m_3 := \lceil m_2(1-\omega) \rceil$ elements of $\hat{X}$ to obtain $X$.
        \State Let us denote the elements of $X$ by $X=[x\ \ y]^T$.
        \State Let $y=ax+b$ be the best fitting (least variance) line to the $(x,y)$ data.
        \For{$k \in \{1,\dots,m_3\}$}
            \State Let $\hat{y}[k]:=ax[k]+b$
        \EndFor
        \State Let $\overline{x}:=$ the mean of $x$ and $\overline{y}:=$ the mean of $y$.
        \State $R^2 = \sum_{k=1}^{m_2} (y[k]-\hat{y}[k])^2 \left/ \ \sum_{k=1}^{m_2} (y[k]-\overline{y})^2 \right.$
        \If{$a \ne 0$ and $m_3 \ge 2$}
            \State $\overline{e}:=\frac{1}{m_3}\sum_{k=1}^{m_2} (y[k]-(ax[k]+b))$
            \State $s^2_e := \frac{1}{m_3-1}\sum_{k=1}^{m_2}(y[k]-(ax[k]+b)-\overline{e})^2)$
            \State Let $t_1,\dots,t_{m_2+v} := p_1,\dots,p_{m_2},q_1,\dots,q_v$
            \State Let us sort $t_1,\dots,t_{m_2+v}$ in an increasing order to obtain $t_{\psi(1)},\dots,$ $t_{\psi(m_2+v)}$.
            \State Let $p_{95} :=$ The $95$'th percentile of the $t_{\psi(1)},\dots,t_{\psi(m_2+v)}$ data.
            \State Let $p_{99} :=$ The $99$'th percentile of the $t_{\psi(1)},\dots,t_{\psi(m_2+v)}$ data.
            \If{$V_{00}V_{11} = 0$}
                \State $r_{\text{corr}} := 0$
            \Else
                \State $r_{\text{corr}} := \frac{V_{01}}{\sqrt{V_{00}V_{11}}}$
            \EndIf
            \State Add the following tuple to the $\mathcal{L}$ output list: $(L,i,R,j,a,b,\overline{e},s^2_e$, $R^2,\mu,V,p_{95},p_{99},r_{\text{corr}})$.
        \EndIf
    \EndIf
\EndFor
\end{algorithmic}
\textbf{Output:} $\mathcal{L}$ trained parameters for the most informative pairs.
\end{breakablealgorithm}

\subsection{The Anomaly Detection test algorithms}

\vspace{3mm}
\begin{breakablealgorithm}
\caption{ECOD Anomaly Detector test}\label{alg:ecod_test}
\textbf{Inputs:} A new $N \in \mathbb{R}^n$ row of variables with indices (dtype,opttype,$j$), and a trained ECOD Anomaly Detector $F_1,\dots,F_n$, $\mu_1,\dots,\mu_n$, $\sigma_1,\dots,\sigma_n$, $s_1,\dots,s_n$, $c,k: \text{ECOD}'$ score parameters, $t$ output cutoff $\in\{0,\dots,n\}$.
\begin{algorithmic}[1]
\State Let $S = [\ ]$ an $n$-long list of $0$s where the anomaly scores will be stored.
\For{$x \in \{N[1],\dots,N[n]\}$}
    \If{$F_j(x)\ne 0$}
        \State $S[j] := -\ln(F_j(x))$
    \ElsIf{$\sigma_j = 0$}
        \State $S[j] := c|x-\mu_j|$
    \Else
        \State $S[j] := \frac{c|x-\mu_j|}{\sigma_j^k}$  (See Formula \ref{eq:ecodprime}.)
    \EndIf
\EndFor
\State Sort $S$ in a decreasing order to obtain the elements $S[\phi(1)],\dots,S[\phi(n)]$.
\State Optionally cut off the first $t$ elements of $S[\phi(1)],\dots,S[\phi(n)]$ and discard the rest to obtain $S[\phi(1)],\dots,S[\phi(t)]$.
\end{algorithmic}
\textbf{Output:} The most anomalistic variables $N[\phi(1)],\dots,N[\phi(t)]$ with anomaly scores $S[\phi(1)],\dots,S[\phi(t)]$.
\end{breakablealgorithm}

\vspace{3mm}
\begin{breakablealgorithm}
\caption{Linear Regression Anomaly Detector test}\label{alg:linreg_test}
\textbf{Inputs:} A new $N \in \mathbb{R}^n$ row of variables with indices (dtype,opttype,$j$), and a trained Linear Regression (and Multivariate Sampling) Anomaly Detector with output list $\mathcal{L}$, $c_A$ Linear Regression A-type anomaly cutoff (by default, $2$ \textit{times the variance}), $c_{AA}$ Linear Regression AA-type anomaly cutoff (by default, $3$ \textit{times the variance}), Linear Regression output cutoff.
\begin{algorithmic}[1]
\State Let $S = [\ ]$ an $|\mathcal{L}|$-long list of $0$s where the anomaly scores will be stored.
\For{$(L,i,R,j,a,b,\overline{e},s_e^2,R^2,\mu,V,p_{95},p_{99},r_{\text{corr}}) \in \mathcal{L}$}
\If{$(L,i)\ne(R,j)$}
    \State Let $x$ be the element of $N$ with index (dtype,opttype,$k$), where (dtype, opttype)=$L$ and $k$=$i$, and let $l:=$(dtype, opttype, $i$).
    \State Let $y$ be the element of $N$ with index (dtype,opttype,$k$), where (dtype, opttype)=$R$ and $k$=$j$, and let $r:=$(dtype, opttype, $j$).
    \State $\hat{y}:=ax+b$
    \State $V_C := (y-\hat{y}-\overline{e})^2$
    \State $A_{\text{score}}:=\frac{1}{\sqrt{2\pi s_e^2}}\exp\left(-\frac{V_C}{2s_e^2}\right)$
    \If{$c_{AA}s_e^2 < V_C$} (See Formula \ref{formula:linreg_saved}.)
        \State $A_{\text{type}} = \text{"AA"}$
    \ElsIf{$c_A s_e^2 < V_C \le c_{AA}s_e^2$}
        \State $A_{\text{type}} = \text{"A"}$
    \Else
        \State $A_{\text{type}} = \text{"B"}$
    \EndIf
    \If{$A_{\text{type}} \ne \text{"B"}$}
        \State Add $(l,r,A_{\text{score}},A_{\text{type}},R^2)$ to $S$.
    \EndIf
\EndIf
\EndFor
\State Sort $S$ based on its $3$rd parameter (the Anomaly Score) in an increasing order.
\State Optionally cut off the last elements of $S$ such that its size becomes the Linear Regression output cutoff.
\end{algorithmic}
    \textbf{Output:} $S$, which is a sorted list of the most anomalistic variable pairs $(l,r)$, with anomaly scores $A_{\text{score}}$, anomaly types $A_{\text{type}}$, and $R^2$ historical correlations..
\vspace{1mm}
\end{breakablealgorithm}

\vspace{3mm}
\begin{breakablealgorithm}
\caption{Multivariate Sampling Anomaly Detector test}\label{alg:mvs_test}
\textbf{Inputs:} A new $N \in \mathbb{R}^n$ row of variables with indices (dtype,opttype,$j$), and a trained (Linear Regression and) Multivariate Sampling Anomaly Detector with output list $\mathcal{L}$, Multivariate Sampling output cutoff.
\begin{algorithmic}[1]
\State Let $S = [\ ]$ an $|\mathcal{L}|$-long list of $0$s where the anomaly scores will be stored.
\For{$(L,i,R,j,a,b,\overline{e},s_e^2,R^2,\mu,V,p_{95},p_{99},r_{\text{corr}}) \in \mathcal{L}$}
\If{$(L,i)\ne(R,j)$}
    \State Let $x$ be the element of $N$ with index (dtype,opttype,$k$), where (dtype, opttype)=$L$ and $k$=$i$, and let $l:=$(dtype, opttype, $i$).
    \State Let $y$ be the element of $N$ with index (dtype,opttype,$k$), where (dtype, opttype)=$R$ and $k$=$j$, and let $r:=$(dtype, opttype, $j$).
    \State $X := [x\ \ y]^T$
    \State $A_{\text{score}} := \frac{1}{\sqrt{(2\pi)^2 \det(V)}} \exp\left(-\frac{1}{2}(X-\mu)^TV^{-1}(X-\mu)\right)$  (See Formula \ref{formula:mvs_saved}.)
    \If{$A_{\text{score}} > p_{99}$}
        \State $A_{\text{type}} := \text{"AA"}$
    \ElsIf{$p_{95} < A_{\text{score}} \le p_{99}$}
        \State $A_{\text{type}} := \text{"A"}$
    \Else
        \State $A_{\text{type}} := \text{"B"}$
    \EndIf
    \If{$A_{\text{type}} \ne \text{"B"}$}
        \State Add $(l,r,A_{\text{score}},A_{\text{type}},r_{\text{corr}})$ to $S$.
    \EndIf
\EndIf
\EndFor
\State Sort $S$ based on its $3$rd parameter (the Anomaly Score) in an increasing order.
\State Optionally cut off the last elements of $S$ such that its size becomes the Multivariate Sampling output cutoff.
\end{algorithmic}
    \textbf{Output:} $S$, which is a sorted list of the most anomalistic variable pairs $(l,r)$, with anomaly scores $A_{\text{score}}$ anomaly types $A_{\text{type}}$, and $r_{\text{corr}}$ historical correlations.
\vspace{1mm}
\end{breakablealgorithm}

\subsection{Algorithm complexities}

\textbf{Complexity of Algorithm \ref{alg:ecod_train}, ECOD Anomaly Detector training}

Omitting all NaN values can be done by going through the entire $m \times n$ table in $O(mn)$ time. The empirical cumulative distribution function is technically a sorted list of all column values. This sorting can be done in $O(mn\log(m))$ time. Later when looking up a value $x$ in list $F$ (namely, calculating $F(x)$), one can use binary search in the sorted list to obtain $F(x)$ in $O(\log(m))$ time. Assuming squaring and cubing a variable takes $O(1)$ time, then calculating all standard deviations and skewnesses takes $O(mn)$ time. Overall, the time complexity of Algorithm \ref{alg:ecod_train}, ECOD Anomaly Detector training is $O(mn\log(m))$.

\vspace{2mm}
\textbf{Complexity of Algorithm \ref{alg:pairs}, Finding the most informative pairs}

Let us denote the number of expert-defined $P=\ $((dtype$_1$,opttype$_1$), (dtype$_2$,opttype$_2$)) pairs by $p$. For each pair $P$, the size of $D_L$ and $D_R$ are $n_L$ and  $n_R$, respectively. Later we will sum up the total complexity for these pairs.

For one given pair $P$, counting the number of missing values takes $O((n_L+n_R)m)$ time, and assuming taking the square root is an $O(1)$ complexity operation, calculating all $p_L(i)$ and $p_R(j)$ values also takes $O((n_L+n_R)m)$ time. Filling out the missing values with the mean of each column takes $O((n_L+n_R)m)$ time, and calculating all standard deviations then removing them also takes $O((n_L+n_R)m)$ time.

Calculating one Kendall-correlations takes $O(m^2)$ time, which we need to find for all combinations of $(i,j), i \in \{1,\dots,n_L\}, j\in \{1,\dots,n_R\}$, which takes a total time of $O(m^2n_Ln_R)$. Penalizing each element of the correlation matrix, and taking the absolute value of it takes a further $O(n_Ln_R)$ time.

Graph $G$ has $n_L+n_R$ vertices and $n_Ln_R$ edges. Calculating its maximum weight spanning tree $T$ takes $O((n_L+n_R)^2)$ time. Discarding all edges with low Kendall-correlation and adding the remaining to $\mathcal{M}$ takes a further $O(n_Ln_R)$ time.

In total, for a single pair $P$, the algorithm takes $O((n_L+n_R)m)+O(m^2n_Ln_R)+O((n_L+n_R)^2)=O(m^2(n_L+n_R)^2)$ time.

We are doing this calculation $p$ times, therefore, the total complexity can be overestimated as $O(pm^2n^2)$, since $n \ge n_L$ and $n \ge n_R$.

However, since there are $10$ different possible dtypes and $6$ different possible opttypes, $p \le 6 \cdot 10 = 60$. Moreover, on average, the size of $n_L$ and $n_R$ are $n/60$. Therefore, $(n_L+n_R)^2 \approx n^2/900$. Because of this, $p(n_L+n_R)^2 \approx pn^2/900 \approx n^2/15$.

Therefore, a better estimate for the total complexity of Algorithm \ref{alg:pairs}, Finding the most informative pairs, is $O(m^2n^2)$, and even then, one should note that the coefficient hidden by $O$ is significantly less than $1$.

\vspace{2mm}
\textbf{Complexity of Algorithm \ref{alg:linreg_mvs_train}, Linear Regression and Multivariate Sampling Anomaly Detector training}

Algorithm \ref{alg:linreg_mvs_train} calls Algorithm \ref{alg:pairs} as a subroutine, therefore we immediately start off with a complexity of $O(m^2n^2)$.

Each expert-defined pair has some number of elements, say $n_L+n_R$. Then the spanning tree in Algorithm \ref{alg:pairs} kept $O(n_L+n_R)$ edges. Therefore, for each of the $p$ expert-defined pairs, we can expect at most $O(n_L+n_R)$ informative pairs found by Algorithm \ref{alg:pairs}, which in total is $O(p(n_L+n_R))$ pairs. However, $n_L$ and $n_R$ are not constants and their size depends on the given pair they represent. Therefore, using the argument at the end of calculating the complexity of Algorithm \ref{alg:pairs}, we can approximate $p(n_L+n_R) \approx pn/30 \le 2n = O(n)$. Therefore, $|\mathcal{M}| = O(n)$. Note that this is an overestimate, since in practice, a lot of pairs will be discarded whenever their Kendall-correlation is low. 

Now net us calculate the time complexity of training the Anomaly Detector for $1$ pair, then at the end, we will multiply the resulting complexity by $O(n)$, since that is how many pairs there are. In order to not underestimate the final time complexity, we will assume that $m_3 \approx m_2 = m$, namely that we won't discard any elements and the final cutoff is also insignificant. Similarly, we will assume both $l$ and $r$ have a relative standard deviation that is larger than the Relative standard deviation cutoff.

For each pair $(l,r)$, calculating their mean, (relative) standard deviation and covariance matrix takes a total of $O(m)$ time. Since $V \in \mathbb{R}^{2\times 2}$, calculating its determinant and inverse take $O(1)$ time. Then calculating the $p_1,\dots,p_m$ values takes $O(m)$ time, and calculating $q_1,\dots,q_v$ takes $O(v)$ time. Sorting $p_1,\dots,p_m$ takes $O(m\log(m))$ time. The cutoffs take $O(1)$ time.

Calculating the best fitting $y=ax+b$ line with singular-value decomposition takes $O(mk)$ time, where in our case, $k=2$, which results in a total of $O(m)$ time. Calculating the averages $\overline{x}$, $\overline{y}$, $R^2$ score, error term $\overline{e}$, and error variance $s_e^2$ all take $O(m)$ time.

Sorting $t_1,\dots,t_{m_2+v}$ takes $O((m_2+v)\log(m_2+v))=O((m+v)\log(m+v))$ time. Afterwards, calculating its percentiles take $O(1)$ time. Finally, calculating $r_{\text{corr}}$ also takes $O(1)$ time.

In total, considering all $O(n)$ pairs, the time complexity of Algorithm \ref{alg:linreg_mvs_train}, Linear Regression and Multivariate Sampling Anomaly Detector training is $O(m^2n^2+n(m+v)\log(m+v))$, which can be simplified to $O(n^2c^2)$, where $c=\max(m,\sqrt{v\log(v)})$. Since usually we have more data than the square root of the sampled values in Multivariate Sampling, the total complexity will usually simplify to $O(n^2m^2)$.

\vspace{2mm}
\textbf{Complexity of Algorithm \ref{alg:ecod_test}, ECOD Anomaly Detector test}

For an input vector of $N \in \mathbb{R}^n$, we need to substitute each element $x$ into the saved $F$ cumulative distribution function, which is an ordered list, so calculating $F(x)$ takes $O(\log(m))$ time. Then calculating the anomaly score takes $O(1)$ time. Therefore, the complexity of Algorithm \ref{alg:ecod_test}, ECOD Anomaly Detector test is $O(n\log(m))$.

\vspace{2mm}
\textbf{Complexity of Algorithm \ref{alg:linreg_test}, Linear Regression Anomaly Detector test}

We will assume that no $(L,i)$ is equal to $(R,j)$ and that all vectors $x$ and $y$ are of maximal length $m$, in order to not underestimate the complexity of this algorithm.

Calculating $\hat{y}$, $V_C$, $A_{\text{score}}$, and finding $A_{\text{type}}$ with the saved values take $O(1)$ time for each pair.

Like we have seen in the complexity calculation of Algorithm \ref{alg:linreg_mvs_train}, there are $O(n)$ informative pairs, therefore, $S$ will be length $O(n)$. Finally, we sort $S$ in $O(n\log(n))$ time.

So the total complexity of Algorithm \ref{alg:linreg_test}, Linear Regression Anomaly Detector test is $O(n\log(n))$.

\vspace{2mm}
\textbf{Complexity of Algorithm \ref{alg:mvs_test}, Multivariate Sampling Anomaly Detector test}

Similarly to before, we will assume that no $(L,i)$ is equal to $(R,j)$ and that all vectors $x$ and $y$ are of maximal length $m$, in order to not underestimate the complexity of this algorithm.

Calculating $X, A_{\text{score}}, A_{\text{type}}$ with the saved values take $O(1)$ time.

Once again, there are $O(n)$ informative pairs, therefore, $S$ will be length $O(n)$. Finally, we sort $S$ in $O(n\log(n))$ time.

So the total complexity of Algorithm \ref{alg:mvs_test}, Linear Regression Anomaly Detector test is $O(n\log(n))$.

\vspace{2mm}
\textbf{Total train and test complexities of the Anomaly Detection methodology}

Summing up the total train complexities we obtain that the Anomaly Detection training algorithms take a total of $O(n^2m^2)$ time (unless for the Multivariate Sampling size parameter $v$, $v\log(v)\ge m^2$, in which case, this complexity becomes $O(n^2v\log(v))$).

And for the Anomaly Detection test algorithms, we obtain a total time complexity of only $O(n\log(n+m))$, which demonstrates that it was worth applying the train-test methodology to fit their Anomaly Detectors and precalculate the commonly used values.

\section{Illustrative Results}\label{section:results}

In this section, we will showcase some of the Anomaly Detection findings. Due to the data being sensitive, we will multiply all numerical values by random number between $[0.9,1.1]$. We will present three examples of anomalies we have found.

In general, the LP solver outputs a result, which, among other data, contains the current plan's Marginal Values. Sorting them in a decreasing order of their absolute value, allows us to filter out the particularly outstanding values of the current plan. However, this does not consider the historical values of them. The Anomaly Detection methodology does use this historical information, and sorts Marginal Values by an Anomaly Score, which indicates how different they are from the historical distribution. 

If the current plan contains an order of magnitude jump Marginal Value, that will be noticed. Marginal Values are usually of varying order of magnitude. If such an order of magnitude jump occurs for typically small Marginal Values, those go unnoticed. 

Large Marginal Values of a constraint may indicate "mathematical strain". This means that if we just had 1 more unit of material to satisfy the constraint, then our profit would increase in accordance. This is also called a "tight" system.

Our goal, in all cases, is first of all to correct all unrealistic values and typos; and secondly, to gain a better business understanding and a trust in the LP solution. The planners need to be confident that their plan is without issues when handing it over to the scheduler colleagues.

\vspace{2mm}
\noindent\textbf{Example 1}
\vspace{2mm}

This is an example of such a large Marginal Value. A Blend constraint in one of the plans had a Marginal Value of $80143.31$. This was significant, because of the $5004$ cases that we trained on, $3501$ contained an entry for this Marginal Value. However, historically, this Marginal Value was always in the range $[-1.02,39.41]$, and most of the time it was below $2$.

This was an internal material flow, whose Marginal Value is expected to be $0$, or at least, very close to $0$, which historically held.

A price of a barrel of oil is in the range of $\$50-\$80$. The barrel factor is around $7.5-8$. This ratio compares the price of products sold that were produced from the given barrel to the purchase cost of the barrel of material. This means that the materials bought for the above $\$50-\$80$, after chemical transformations, are sold for around $\$350-\$700$. This does not mean that the profit is $\$300-\$600$, because the cost of refining takes up the majority of this price. In fact, a healthy margin on a given barrel is around $\$3-\$4$.

Marginal Values represent a similar idea: they indicate how much profit we could generate by adding infinitesimally more input material. The value of this infinitesimally small material should realistically not be greater than the margin on a barrel of material.

However, in this case, the obtained Marginal Value was much larger than expected, $80143.31$. The potential causes of this may be one or more of the following:

\begin{enumerate}
\item The output of the LP was almost infeasible.
\item The constraint corresponding to this Marginal Value is "tight".
\item Tight specifications on the recipe in the current situation.
\item The system cannot use up the materials included in this blend.
\item etc.
\end{enumerate}

The potential solutions for $1-4$ are the following:

\begin{enumerate}
\item[1-2.] Along with other large Marginal Values, find related capacities / maximum values that are possible to increase to "loosen" the system.
\item[3.] If the specification is tight, release it.
\item[4.] Find material balance limits that can be released. Or provide the input materials from other inventories as well.
\end{enumerate}

\begin{figure}[h]
    \centering
    \includegraphics[width=0.8\linewidth]{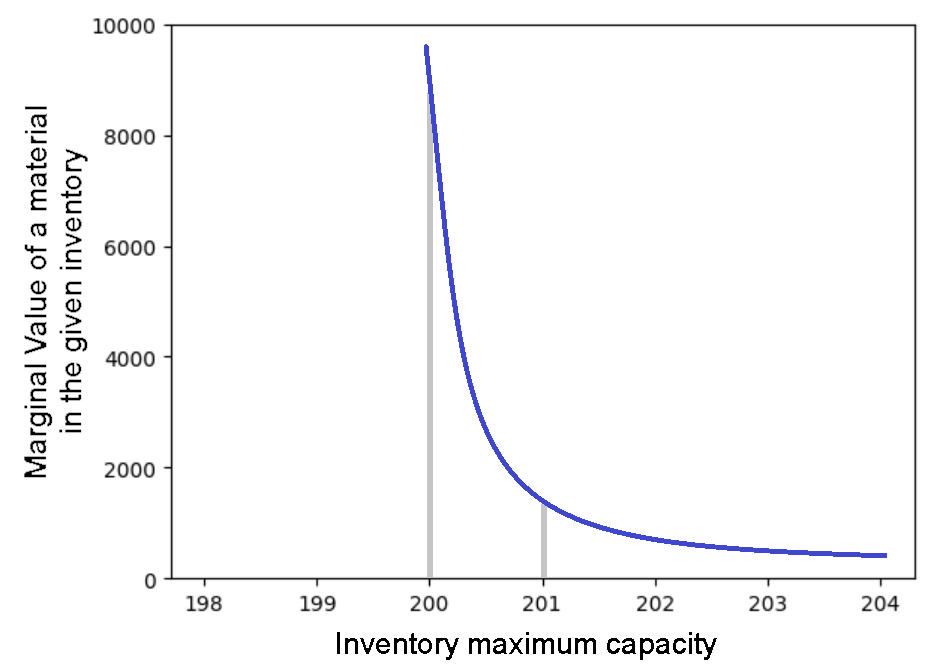}
    \caption{If the given inventory maximum capacity is increased by $1$ ("loosening" the model), the corresponding Marginal Value may decrease to realistic values. The magnitude of the jump depends on the magnitude of the material balance value. Of course, we do not know the shape of the above curve, we can only find its values by reevaluating the model.}
    \label{fig:inventory_margval_relation}
\end{figure}

An example for altering the amount of material used for this blend from a given inventory is to increase it from by about $1$ t, as seen on Figure \ref{fig:inventory_margval_relation}. This requires modifying other constraints as well, since materials from the same inventory may be used elsewhere. Compared to the total material balance, this change is minor, but mathematically it releases the constraint, obtaining a realistic Marginal Value.

Through this process, we can iterate the model to find realistic values, in order to hand over a trustworthy LP solution.

\vspace{2mm}
\noindent\textbf{Example 2}
\vspace{2mm}

This is an example of a pair-anomaly we found. We are doing a weight-proportional blending of several materials. Two components of this blend, component proportions $A$ and $B$ used to historically always have a relation of $B = 1.5133\cdot A$, however, in the currently tested case, $B=0.8787\cdot A$. This is why the Linear Regression Amomaly Detector output this pair as an anomaly. Moreover, the Multivariate Sampling Anomaly Detector also found this to be an anomaly, because historically both $A$ and  $B$ were small, below $0.08$, but in the currently tested case, both, $A$ and $B$ were above $0.3$. Therefore, this pair is in the orange zone of Figure \ref{fig:anomaly_regions}, so it is a Significant Anomaly.

There are two types of blending: formula blend, and specification blend. Formula blends always follow the same proportions in every blend. In the case of specification blends, the system can set these proportions (between given limits), with the goal of outputting a blend with a given quality.

Whenever experts find such an anomaly, they need to check which type of blend this was. This was a formula blend, so planners need to be aware of the change. In this case, if this showed up as an anomaly, we need to disregard it, since this originated from a model change.

In case of a specification blend, it is possible that the external environment changes, e.g. the price structure is different; or that the internal environment changes, e.g. which plants are under maintenance, what kinds of materials are available. In this case, achieving the same quality may require unusual blend proportions. If we believe that these were not causing the issue, then data supply for the limits or capacities can be incorrect.

If the model ran without infeasibility, and with correct data supply, then it was capable of fulfilling the quality constraints. So in this case, even if this blend is unusual compared to historical blends, then it should not cause any issues.

\vspace{2mm}
\noindent\textbf{Example 3}
\vspace{2mm}

This is an example of two inventory capacity maximums, $A$ and $B$. Historically we have found that $B = 0.5916\cdot A-0.6159$ holds with an $R^2 = 0.8225$. However in this case, $A$ and $B$ were extremely far away from this line.

The potential causes for this anomaly may be data supply issues, or an actual inventory capacity increase or decrease. Sometimes, due to operational requirements, schedulers can add or remove extra tanks, which could be the cause of this anomaly.

If this was not an operative change, then experts will need to discuss with the colleagues responsible for this inventory whether or not this was an actual inventory capacity increase (physically more barrels are available or additional warehouses are rented), or if it was an error.

\vspace{2mm}
\noindent\textbf{Example 4}
\vspace{2mm}

This example is, as of yet one of the best Anomaly Detector findings. There exist domestic and EU regulations for all output materials (gasoline, kerosene, etc.) to have extremely low sulfur amounts (at most 10 ppm). As such, there is a maximum constraint on the sulfur concentration of most internal flows. One internal flow seemed to be struggling to keep up with its sulfur constraint.

The Anomaly Detector has output a high Marginal Value on this plant, around $3-4$ \$/ppm. While historically the same value was only around $0-0.8$ \$/ppm. Even though this is over a threefold increase, according to the planning experts, this is a significant error that would have slipped through the cracks due to the hundreds of thousands of non-zero values used in each plan.

The cause of this issue turned out to be a measurement / simulation error on the sulfur concentration of a plant that provided one of the inputs to to this plant. This relatively small measurement error was not caught, and it caused the whole system to drastically alter its blending proportions for a number of products.

Once we caught the error and the system recognized the correct sulfur concentrations, it automatically returned to the previous, less tight blend proportions.

\section{Summary}\label{section:summary}

In this paper, we introduce a methodology that enhances the capabilities of linear programming solvers, particularly when dealing with large-scale optimization problems, by incorporating historical information. This approach enables the identification of unusual values and relationships that can provide decision-makers with additional insights beyond those offered by conventional solver outputs. Such information cannot be derived from the optimization results alone; rather, it emerges when the historical context of the problem is also taken into account. This, however, introduces a further challenge: identifying unusual patterns within large datasets. To make this unsupervized learning tractable we introduced an Anomaly Detection tool.

Data gathered from various sources, such as external datasets, internal records, or software may overlook key insights relevant to experts. Moreover, refinery optimization frequently simplifies the model (such as planning for a singular point in time instead of several points for one month), which often results in several unrealistic outputs, when handled without care.
This new approach is not only capable of finding unusual, unrealistic and unattainable values that hinder planning, but also uncover hidden potentials and bottlenecks in each plan. By using not only the input and output values of the LP problem, but also each constraint's Marginal Value, the found anomalies may indicate new business opportunities that could not have been considered without Anomaly Detection due to the sheer size of the data.

The first step in dimensionality reduction involves defining different levels at which the anomaly analysis is performed separately using the introduced Anomaly Detection tool. These levels are determined based on expert knowledge.

The Anomaly Detection methodology developed splits the problem into two sub-problems: one- and multidimensional Anomaly Detection. For the one-dimensional case, we have applied and modified the well-known ECOD methodology to fit our problem. For the multidimensional case, we found the most informative pairs (a method which reduces the large number of two dimensional margins massively).


The multivariate Anomaly Detector tool contains first an automatic Kendall-correlation and spanning tree-based filtering.  We have introduced two new methodologies for finding 2D anomalies in this filtered dataset. Using historical data, we have trained a Linear Regression-based and a Multivariate Sampling-based algorithm, which can then be used to detect 3 types of 2D anomalies. By choosing fitting names for these groups, they can then be easily interpreted by experts.

Using these introduced methods, we have found several 1D and 2D anomalies. Due to the sheer amount of data available, it would be impossible to manually find meaningful anomalies. Once the Anomaly Detector found the largest anomalies, and sorted by Anomaly Score, then experts can analyze at them, and using their experience, they can interpret these results with improved business understanding.

We believe that this work opens up new avenues for gaining valuable insights by integrating the proposed Anomaly Detector into decision making based on LP solvers that are executed periodically. By leveraging information from previous runs, this approach makes it possible to identify unusual patterns and uncover potential business opportunities hidden within Big Data.

In the future, one may extend this methodology into finding 3D anomalies, and making them interpretable. Alternatively, one may consider using distributions other than the Gaussian distribution.

\newpage
\vspace{3mm}
\noindent {\large\bfseries Acknowledgements}
\vspace{3mm}

The research reported in this paper is part of project no. BME-NVA-02, implemented with the support provided by the Ministry of Innovation and Technology of Hungary from the National Research, Development and Innovation Fund, financed under the TKP2021 funding scheme. Additional funding was provided by NKFIH for the Cooperative Doctoral Program.


\newpage
\bibliographystyle{elsarticle-num} 



\end{document}